\documentclass[lettersize,journal]{IEEEtran}
\usepackage{amsmath,amsfonts}
\usepackage{algorithmic}
\usepackage{algorithm}
\usepackage{array}
\usepackage[caption=false,font=normalsize,labelfont=sf,textfont=sf]{subfig}
\usepackage{textcomp}
\usepackage{url}
\usepackage{verbatim}
\usepackage{graphicx}
\usepackage{cite}

\usepackage[numbers]{natbib}
\usepackage[cmintegrals]{newtxmath}
\usepackage{booktabs}
\usepackage{bm}
\usepackage{newtxtext,newtxmath}
\usepackage{dblfloatfix}
\usepackage{hyperref}

\usepackage{etoolbox}
\begin{document}

\title{Dynamic Avatar-Scene Rendering from Human-centric Context}

\author{Wenqing~Wang,
        Haosen~Yang,
        Josef Kittler,
        Xiatian~Zhu
        
        % <-this % stops a space
\thanks{Wenqing Wang, Josef Kittler are with the Centre for Vision, Speech and Signal Processing (CVSSP), University of Surrey. (e-mail: wenqing.wang@surrey.ac.uk, j.kittler@surrey.ac.uk). Haosen Yang and Xiatian Zhu are with Centre for Vision, Speech and Signal Processing (CVSSP) and People-Centred Artificial Intelligence (PAI), University of Surrey. (e-mail: h.yang@surrey.ac.uk, xiatian.zhu@surrey.ac.uk)}
}

        % <-this % stops a space



\maketitle
\begin{abstract}
Reconstructing dynamic humans interacting with real-world environments from monocular videos is an important and challenging task. 
Despite considerable progress in 4D neural rendering, existing approaches either model dynamic scenes holistically or model scenes and backgrounds separately aim to introduce parametric human priors.
However, these approaches either neglect distinct motion characteristics of various components in scene especially human, leading to incomplete reconstructions, or ignore the information exchange between the separately modeled components, resulting in spatial inconsistencies and visual artifacts at human-scene boundaries.
To address this, we propose {\bf Separate-then-Map} (StM) strategy that introduces a dedicated information mapping mechanism to bridge separately defined and optimized models.
Our method employs a shared transformation function for each Gaussian attribute to unify separately modeled components, enhancing computational efficiency by avoiding exhaustive pairwise interactions while ensuring spatial and visual coherence between humans and their surroundings.
Extensive experiments on monocular video datasets demonstrate that StM significantly outperforms existing state-of-the-art methods in both visual quality and rendering accuracy, particularly at challenging human-scene interaction boundaries. We provide demo videos in supplementary and the code will be released upon acceptance.

\end{abstract}

\begin{IEEEkeywords}
4D reconstruction and animation, digital human avatar, 4D gaussian splatting.
\end{IEEEkeywords}

\section{Introduction}
\label{sec:intro}
% main weakness from reviewers:

% 1. depth (the depth can already solve the problem because provide geometry guidance) - show the ablation about if estimated depth not as 

% 2. prove (do not have explicit prove to show the misalignment between the scene model and the human avatar model) - change the representation 'misalignment' to geometry unalignment

% 3. representation (alignment not exchange information between the background and foreground) - project to the same space

% \begin{figure}[t]
%   \centering
%   % \fbox{\rule{0pt}{2in} \rule{0.9\linewidth}{0pt}}
%    \includegraphics[width=1\linewidth]{ICCV2025-Author-Kit-Feb/images/fig1.2.pdf}

%    \caption{Learning from a limited monocular video of a human moving around the scene, existing approaches face several challenges: (a) Unified 4D reconstruction methods \cite{qian20243dgs} struggle to maintain the integrity of the avatar. (b) Separate-based methods \cite{kocabas2024hugs} suffer from unexpected occlusions and floating artifacts because of the geometric inaccuracies in the interaction regions between the background and the avatar. (c) In contrast, our StP method achieves more accurate and consistent reconstruction by effectively addressing these limitations.
%    %         our StA method allows to render and animate the human avatar in novel poses, in novel views, and even in novel scenes with substantially superior fidelity and realism, compared to the current state of the art HUGS \cite{kocabas2024hugs}.
% }
%    \label{fig:overall}
% \end{figure}

\begin{figure*}[t]
  \centering
   \includegraphics[width=1\linewidth]{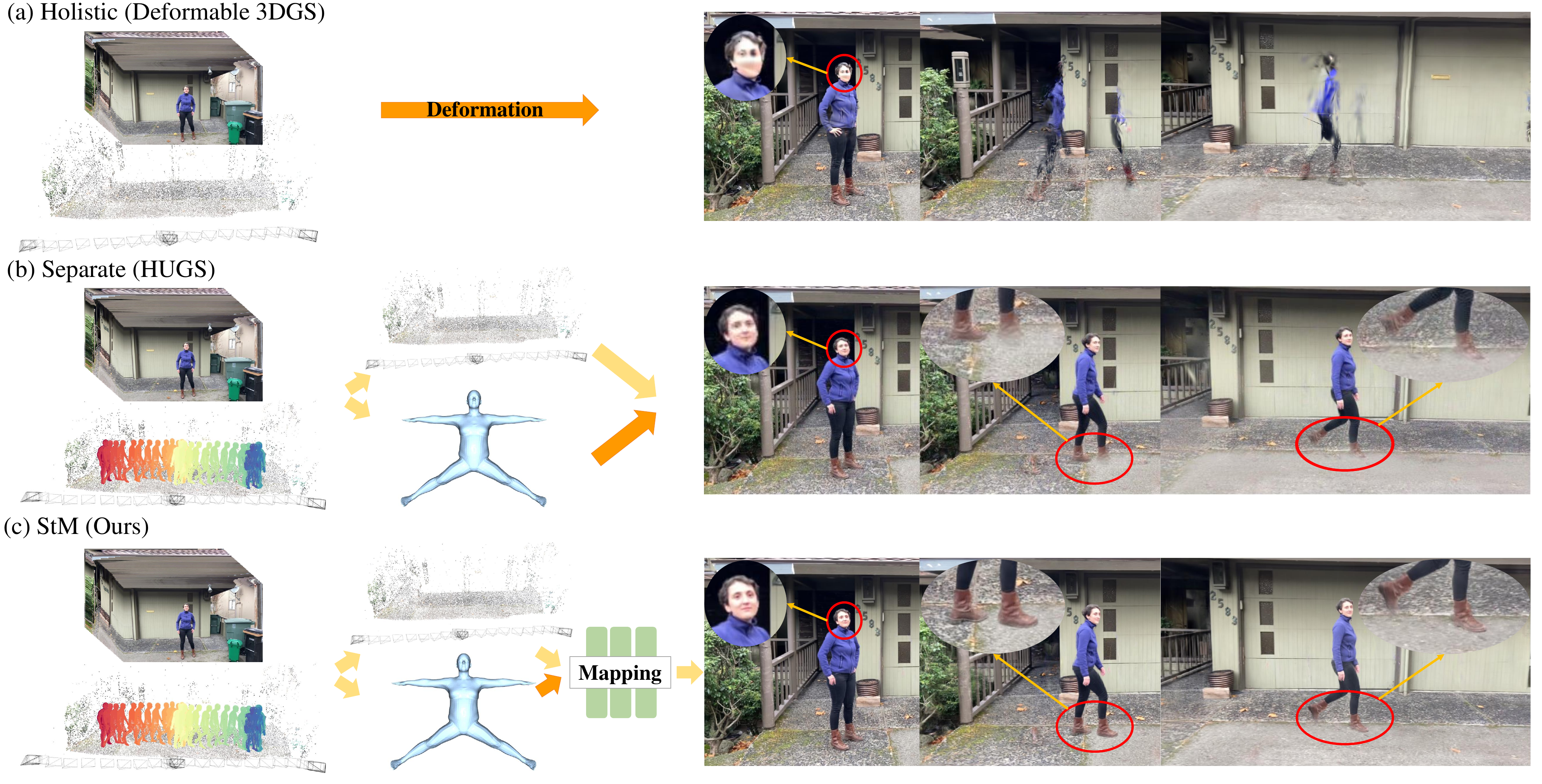}

   \caption{
    Learning from a limited monocular video of a human moving around the scene, existing approaches face several challenges: (a) Holistic 4D reconstruction methods \cite{yang2023deformable3dgs} struggle to maintain the integrity of the avatar. (b) Separate-based methods \cite{kocabas2024hugs} suffer from unexpected occlusions and floating artifacts in the regions where the background and the avatar interact. (c) In contrast, our Separate-then-Map (\textbf{StM}) strategy achieves more accurate and complete reconstruction by mapping different model representations to a unified space.
    }
   \label{fig:overall}
\end{figure*}

\IEEEPARstart{H}{uman-centric} monocular videos, which capture people interacting with real-world environments, are a common data source for applications such as AR/VR, visual effects, gaming, and simulation. While much progress has been made in reconstructing isolated human avatars in clean backgrounds \cite{weng2022humannerf, hu2024gauhuman, hu2024gaussianavatar, moon2024expressive, wang2024single, qian20243dgs, hu2025tgavatar}, relatively less attention has been given to rendering humans together with their surrounding environments, which is essential for modeling realistic human-scene interactions.

Recent works \cite{du2021nerflow, liu2023robust, Wu_2024_CVPR, yang2023deformable3dgs, kocabas2024hugs, jiang2022neuman, guo2023vid2avatar} attempt to reconstruct human-centric scenes by jointly modeling the human and background.
These approaches can be broadly classified into two distinct methodologies.
One line of work, represented by holistic 4D scene modeling approaches~\cite{du2021nerflow, Wu_2024_CVPR, yang2023deformable3dgs, li2025frpgs, guo2024motion}, employs a unified 4D representation to model the entire scene, thereby requiring the simultaneous capture of drastically different motion patterns, specifically, fast-moving humans with complex articulated poses and static or slowly-changing backgrounds.
This inherent conflict in motion characteristics impedes effective disentanglement and reconstruction of foreground and background components, particularly under the large motions, complex poses, and limited viewpoints typical of monocular videos~\cite{qian20243dgs}.
As evident in Figure~\ref{fig:overall}(a), this holistic strategy leads to instability for dynamic object reconstruction.

Alternatively, a more effective approach explicitly separates the human and background into independently modeled components, as demonstrated in \cite{liu2023robust, jiang2022neuman, guo2023vid2avatar, kocabas2024hugs}. This decomposition benefits from human-specific priors (e.g., SMPL \cite{SMPL:2015}, FLAME \cite{FLAME:SiggraphAsia2017}), which provide strong constraints for accurately modeling human motion and articulation, but also introduces new challenges when the separately reconstructed components are combined.
For example, recent work \cite{kocabas2024hugs} combining 3D Gaussian Splatting (3DGS) with parametric-based human-deformable Gaussian Splatting to represent the scene and human avatar separately, ultimately merging the two independent models through direct concatenation.
This type of direct merging strategy introduces floating artifacts and partial occlusions at the interaction boundaries (Figure~\ref{fig:overall}(b)). These issues arise due to separate initialization and differing design and optimization strategies, which hinder information integration between the two components.

To overcome this largely overlooked issue, we propose { \bf Separate-then-Map} (StM) strategy for human-centric 4D scene reconstruction, as shown in Figure~\ref{fig:overall}(c). Unlike previous separate-based methods that directly concatenate independently optimized components, our approach introduces a dedicated information mapping mechanism that projects both the foreground (human) and background models into a unified representation space. This simple yet effective design explicitly addresses the fundamental challenge of integrating models with different initialization strategies, architectural designs, and optimization objectives.
The key insight is that without proper alignment, these separately modeled components, despite being concatenated during rendering, lack a unified representation space, leading to spatial inconsistencies, floating artifacts, and incorrect occlusions, particularly at human-scene interaction boundaries.

However, establishing explicit communication between these components is challenging because both Gaussian fields contain millions of Gaussian primitives with fundamentally different spatial structures.
To address this, we introduce a shared non-linear transformation function that operates across both sets of Gaussian fields. 
This shared design maximizes computational efficiency by avoiding exhaustive pairwise interactions between all foreground and background primitives.
Instead, it utilizes learnable projections along the attribute dimension to map both Gaussian fields into a unified representation space.
Specifically, we apply separate transformations for each Gaussian attribute, including position, color, opacity, rotation, and scale, which enabling attribute-specific alignment. This per-attribute processing ensures that each property is aligned according to its distinct semantic role in the rendering process.

This work makes three contributions:
(1) We systematically identify and investigate a previously overlooked problem of missing information communicate between separately modeled foreground and background in human-centric 4D scene reconstruction.
(2) We propose StM, which introduces a shared information mapping mechanism to project separately initialized, designed, and optimized components into a unified space, ensuring both computational efficiency and spatial coherence.
(3) Extensive experiments on monocular video datasets demonstrate that StM significantly outperforms existing state-of-the-art methods in both visual quality and rendering accuracy.

\section{Related Work}
\label{sec:related_work}
\subsection{4D novel view synthesis and reconstruction} 
4D Novel view synthesis and reconstruction provides a general and holistic solution for reconstructing dynamic objects and scene contents. Recent methods have introduced various dynamic representations such as dynamic NeRF \cite{li2022neural, li2021neural, park2021hypernerf}, dynamic triplane \cite{fridovich2023k, cao2023hexplane}, and dynamic Gaussian Splatting \cite{yang2023real, yang2023deformable3dgs, Wu_2024_CVPR, luiten2023dynamic, bae2024per}, enabling high-quality rendering from both calibrated multi-view and uncalibrated monocular RGB video inputs.
Especially, point-based explicit representations (e.g., Gaussian Splatting) have achieved state-of-the-art rendering quality while maintaining real-time inference and training efficiency.
Dynamic 3DGS \cite{luiten2023dynamic} optimize the position and shape of each Gaussian kernel frame-by-frame, while Deformable 3DGS \cite{yang2023deformable3dgs} and 4DGS \cite{Wu_2024_CVPR} introduce time-dependent deformation networks to deform a canonical 3D Gaussian into each frame.
\cite{yang2023real, li2024spacetime} expand 3D Gaussian primitives to 4D by incorporating temporal properties.
Despite their effectiveness, these approaches are most suitable for scenes with slow-moving objects \cite{park2021hypernerf, li2022neural}. 
Furthermore, they treat the entire scene holistically without integrating statistical models for regularization. This limitation is particularly significant in human-centric scenarios, where prior body models such as SMPL \cite{SMPL:2015} can provide valuable structural constraints. Therefore, our method incorporates human prior knowledge to better handle large motions.

\begin{figure*}[htbp]
  \centering
   \includegraphics[width=1\linewidth]{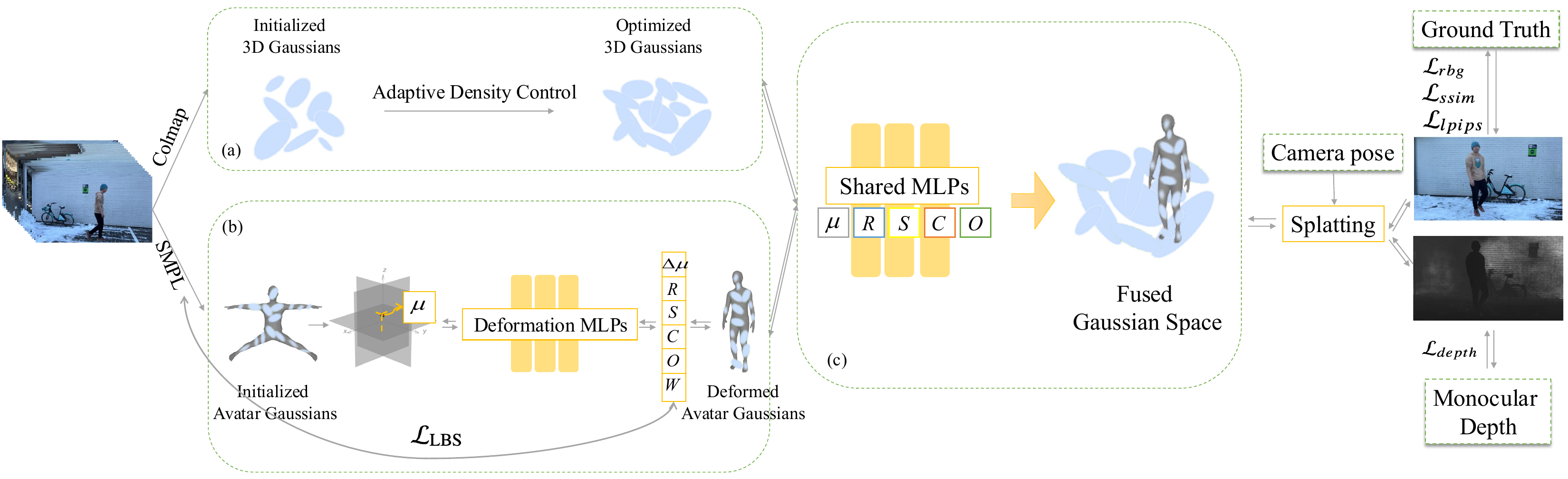}

   \caption{
\textbf{Overview of Separate-then-Map (\textbf{StM})}: Given an input video sequence, we first initialize the point clouds for the scene and avatar Gaussians using Colmap predictions and SMPL vertex points. This decoupled design is as follows: (a) a 3D Gaussian Splatting (3DGS) model represents the background scene, (b) a deformable 3D Gaussian avatar model driven by linear blend skinning (LBS) to represent the foreground human, with the parameters including position offset \(\Delta \mu\), rotation \(R\), scale \(S\), spherical harmonics (SH) coefficients \(C\), opacity \(O\), and LBS weight \(W\), all predicted from the position triplane feature \(\mu\); (c) A information mapping process is then employed to project the scene model and the avatar model into a consistent space. During training, the rendered images and depth maps are used to 
compute the loss
against the ground truth images and monocular estimated depth maps.
}
   
   % Overview of Separate-then-Align({\bf StA}): Given a video sequence as input, we firstly initial point cloud for scene and avatar Gaussians using Colmap predictions and SMPL vertices points. Seprate Gaussian Spaltting model define as below: (a) a static 3DGS as the scene model to present the background, which will be optimized via average gradient; 
   % (b)  a LBS-driven deformable 3D Gaussians as avatar model to repsent the foreground,  position offset $\Delta \mu $, rotation $R$, scale $S$, SH $C$, opacity $O$ and LBS weight $W$ driven by pridicting from position triplane $\mu $ feature to predict;
   % (c) Subsequently, a alignment process is then employed to project the stactic scene model and deformed avatar model to same consistence representation space; 
   % Finally, rendered image and depth map, can used to calculate photometric loss and depth correlation coefficient loss with ground truth image and monocular estimated depth. }
   \label{fig:pipeline}
\end{figure*}

\subsection{Human avatars reconstruction and animation} 
Human avatars reconstruction and animation from visual observations remains a challenging task.
Traditional methods relying on explicit, predefined parametric mesh topologies \cite{SMPL:2015, li2017learning, blanz2023morphable, romero2022embodied} often struggle to capture personalized appearance details \cite{alldieck2018detailed}.
With the advancements in neural representations \cite{mildenhall2021nerf, kerbl3Dgaussians}, recent studies \cite{jiang2023instantavatar, zhang2025humanref, weng2022humannerf, chen2023fast, chen2021snarf, gafni2021dynamic, yu2023monohuman, peng2021animatable, hu2024gauhuman, hu2024gaussianavatar, moon2024expressive, paudel2024ihuman, shao2024splattingavatar, wen2024gomavatar, qian20243dgs} have integrated parametric template models \cite{SMPL:2015, li2017learning, blanz2023morphable, romero2022embodied} to reconstruct dynamic human avatars. 
These hybrid approaches not only capture fine-grained, personalized details but also enable the animation of reconstructed avatars through simple pose guidance.
% Such hybrid representations capturing detailed personalized details, also allow animating the reconstructed avatar through simple pose guided. 
However, the aforementioned methods primarily focus on reconstructing and animating avatars in isolation, disregarding the surrounding environment.
In real-world scenarios, where visual inputs contain rich contextual information, it is essential to reconstruct both the avatar and its environment to achieve a more comprehensive understanding of the scene. 
Our method aims to reconstruct and animate human avatars within their captured environments.

\subsection{Dynamic human reconstruction and animation in context} 
Dynamic human reconstruction and animation in context is becoming increasingly important as virtual avatars interact more closely with their surrounding environments.
Recent methods \cite{jiang2022neuman, guo2023vid2avatar, kocabas2024hugs} have explored detaching the avatar from the background using separate neural fields while incorporating strong human priors to simplify the overall reconstruction process.
Specifically, Neuman \cite{jiang2022neuman} trains two NeRF models \cite{mildenhall2021nerf} - the scene NeRF and the human NeRF successively - encoding the human's appearance and geometry separately from the background.
Similarly, Vid2Avatar \cite{guo2023vid2avatar} models the foreground human and background scene using two separate neural fields without relying on external segmentation modules. 
Utilizing 3D Gaussian Splatting, HUGS \cite{kocabas2024hugs} defines two sets of Gaussian fields with distinct predefined properties, which are subsequently concatenated and rendered together during splatting and optimization.
Separating the background and avatar is an efficient strategy for reconstructing complex human motion while maintaining photorealistic rendering. 
However, existing methods do not consider how to project and recombine different model representations, potentially leading to inconsistencies in integration. To overcome this limitation, our method maps the representation spaces of the two models into a unified one during training, enabling more coherent and realistic reconstructions.

\section{Method}
\label{sec:method}

\noindent{\bf Problem definition}
Given a monocular video captured by a moving camera that shows a human moving through a natural background, our goal is to learn a 4D model capable of rendering the human from novel viewpoints and animating the human in novel poses, both within the observed scene and in new scenes. Monocular video is typically short, providing limited views and poses of the human while continuously moving around the scene. This setup introduces challenges due to sparse views, significant body displacement, and variation in both the scene and the human over time.

% Given a monocular video captured by a moving camera depicting a moving human in a natural background, we aim to learn a 3D model that can render the human in novel views and animate the human in novel poses in this observed scene and other novel scenes. Often this video comes with a short duration, offering limited human views and poses while continuously moving around the scene.
% This poses challenges in training data size, sparse views, large body movement, as well as the variation in both scene and human body over time.
% which is the most common scene we can get in our daily life, called the Human-Centric Scene. 

\noindent{\bf Model architecture}
To address this challenging human-centric 4D scene reconstruction problem, we propose StM strategy.
Specifically, we adopt a decoupling design for background and foreground, where the scene and human are parameterized independently with specialized representation models. 
After separately modeling, we introduce a shared information mapping module that project information from both Gaussian models into a shared space.
This transformation effectively mitigates issues related to separate initialization, and distinct model structures and optimization processes.
% projection, transformation
Our overall architecture, illustrated in Figure~\ref{fig:pipeline}, consists of:
(a) a scene model, (b) a deformable human avatar model, and (c) a shared information mapping module.

% However, this approach introduces the risk of geometric discontinuity between the scene model and the human avatar model due to their separate representations, complicating coherent scene representation and rendering. To address this, an inter-model projection module is necessary to ensure alignment between the two models. This leads to our architecture, as illustrated in Figure~\ref{fig:pipeline}, consisting of:
% (1) a scene model, (2) a human avatar model, and (3) the inter-model projection module.

% To tackle this challenging scene reconstruction problem,
% we adopt a background and foreground decoupling design,
% where the scene and the human are parameterized individually
% using a specific representation model.
% However, this introduces a potential problem that the scene model and the human avatar model may be finally unaligned due to their separation, making trouble for whole scene representation and rendering.
% Under this consideration, an inter-model alignment module is thus necessary to convert the two models to be aligned.
% This leads to our architecture, as shown in Figure~\ref{fig:pipeline}, consisting of:
% (1) A scene model, (2) a human avatar model, and (3) inter-model alignment.

% \noindent{\bf Scene model}
\subsection{Scene model}

We adopt 3DGS \cite{kerbl3Dgaussians} for scene representation due to its effective balance between efficiency and accuracy. First, camera poses and the initial point cloud are estimated from a given video sequence using Colmap \cite{schonberger2016structure, schonberger2016pixelwise}. The point cloud is initialized as a set of 3D Gaussian primitives, denoted $\mathcal{G}_b$.
Each Gaussian primitive $G_k$ is defined by its position (mean) $\boldsymbol{\mu}$, covariance matrix $\boldsymbol{\Sigma}$ and opacity $o$:
\begin{equation}
\label{Eq:3DGS_represent}
    G_k = exp(-\frac{1}{2}(\boldsymbol{x}-\boldsymbol{\mu})^{T}\boldsymbol{\Sigma}^{-1}(\boldsymbol{x}-\boldsymbol{\mu})).
\end{equation} 
The covariance matrix $\boldsymbol{\Sigma}$ factorized into a scaling matrix $S$ and a rotation matrix $R$: $\boldsymbol{\Sigma} = RSS^{T}R^{T}$.

% We consider 3DGS \cite{kerbl3Dgaussians} for scene representation due to its superior trade-off between efficiency and accuracy. 
% The camera poses and initial point cloud are estimated from video sequence utilize Colmap \cite{schonberger2016structure, schonberger2016pixelwise}.
% The background scene represents as a set of 3D Gaussians, denoted $\mathcal{G}_b$.
% Each Gaussian point $G_k$ is define a position (mean) $\boldsymbol{\mu}$, covariance matrix $\boldsymbol{\Sigma}$ and opacity $o$:
% \begin{equation}
% \label{Eq:3DGS_represent}
%     G_k = exp(-\frac{1}{2}(\boldsymbol{x}-\boldsymbol{\mu})^{T}\boldsymbol{\Sigma}^{-1}(\boldsymbol{x}-\boldsymbol{\mu})).
% \end{equation} 
% And the covariance matrix $\boldsymbol{\Sigma}$ factorized into a scaling matrix $S$ and a rotation matrix $R$: $\boldsymbol{\Sigma} = RSS^{T}R^{T}$.

For rendering, the 3D Gaussians are projected onto the 2D image plane to form 2D Gaussians, denoted $G^{2D}_k$.
For any given camera pose $P_i$, an image can be rendered by splatting and alpha blending $N$ sorted Gaussians visible in this view:
% When rendering, the 3D Gaussians are projected onto 2D image plane to represent 2D Gaussians as $G^{2D}_k$, 
% and given \textit{any} camera pose $P_i$, it can be rendered by performing splatting and alpha blending on $N$ sorted Gaussians visible at this view:
\begin{equation}
    I^{\prime} = R(\mathcal{G}_b, P_{i}) =
    \sum^{N}_{k=1}  o_k G^{2D}_k  c_k(P_{i})
    \prod^{k-1}_{j=1} (1- o_j G^{2D}_j),
    \label{eq:3DGS_rendering}
\end{equation}
where $R$ represents the differentiable Gaussian rasterizer, $c_k$ is the view-dependent color of the $k$-th Gaussian, $I^{\prime}$ denotes the rendered image, which can be used to calculate photometric loss with ground truth images.

\subsection{Human avatar deformation model}
Combine parametric human prior model, such as SMPL \cite{SMPL:2015}, with implicit or explicit neural rendering enables more complete reconstruction and novel pose animation, as demonstrated by several works \cite{weng2022humannerf, jiang2022neuman, lei2024gart, hu2024gauhuman}.
Therefore we utilize a Linear Blend Skinning (LBS)-driven deformable 3D Gaussians $\mathcal{G}_a$ \cite{kocabas2024hugs} to represent the human avatar. 
The estimated SMPL parameters $\bm{\beta}$ and $\bm{\theta}$ for each frame are used to regularize body shapes and poses \cite{SMPL:2015, goel2023humans}.

Specifically, for each Gaussian kernel, taken its position triplane feature as input, which is passed through three MLP deformation modules to predict Gaussian point's position offset, color, opacity, rotation, scale and LBS weights.
The movement of each Gaussian primitive from the canonical position $\bm{p}^c$ to posed position $\bm{p}^\theta$ is driven by the optimized LBS weights using forward skinning:
\begin{equation}
\label{Eq:smpl_lbs}
    \bm{p}^\theta = \sum_{k=1}^{K}w_{k}(\bm{B_k}(\bm{J}, \bm{\theta})\bm{p}^c+\bm{b_k}(\bm{J}, \bm{\theta}, \bm{\beta})),
\end{equation} 
where $\bm{J}$ includes $K$ joint locations, $\bm{B_k}(\bm{J}, \bm{\theta})$ and $\bm{b_k}(\bm{J}, \bm{\theta}, \bm{\beta})$ are the transformation matrix and translation vector of joint $k$ respectively, $w_{k}$ is the linear blend weight for this Gaussian primitive. 

% $\bm{\beta}, \bm{\theta}$ are the body shape and pose parameters, 

\subsection{Human and Scene shared information mapping}
\label{sec:scene_human_align}

The separate modeling of the scene and human avatar have suboptimal initialization and employ different model design and optimization strategies, although they are concatenate together for rendering and gradient back-propagation in the previous method \cite{kocabas2024hugs}, still no direct correspondence or pairing relationship between scene and avatar Gaussian primitives, which will cause conflict and geometry inconsistency, displaying outside as floating artifacts and foreground occlusions.
To map separately modeled components into an unified space, we introduce a shared information mapping module for per-attribute.

\begin{table*}[t]
    \centering
    \caption{Comparing the novel view synthesis results of the whole {\bf\em scenes} on Neuman \cite{jiang2022neuman} dataset. 
    The front part is holistic 4D scene reconstruction methods, and the second part is separate modeled reconstruction methods.
    And the last part is our proposed StM results.
    Note, 4DGS failed to converge on the Lab scene, resulting in no results.
    HUGS* is our re-implementation trained for 20,000 iterations compared with original 15,000 iterations.
    % We outperform all competitive baselines in all three metrics.
    }
    \setlength{\tabcolsep}{3pt}
    \resizebox{\textwidth}{!}{
    \begin{tabular}{l|ccc|ccc|ccc|ccc|ccc|ccc}
    \toprule
        & \multicolumn{3}{c|}{\textbf{Seattle}} & \multicolumn{3}{c|}{\textbf{Citron}} & \multicolumn{3}{c|}{\textbf{Parking}} & \multicolumn{3}{c|}{\textbf{Bike}} & \multicolumn{3}{c|}{\textbf{Jogging}} & \multicolumn{3}{c}{\textbf{Lab}}   \\
    \midrule
        & PSNR$\uparrow$ & SSIM$\uparrow$ & LPIPS$\downarrow$ & PSNR$\uparrow$ & SSIM$\uparrow$ & LPIPS$\downarrow$ & PSNR$\uparrow$ & SSIM$\uparrow$ & LPIPS$\downarrow$ & PSNR$\uparrow$ & SSIM$\uparrow$ & LPIPS$\downarrow$ & PSNR$\uparrow$ & SSIM$\uparrow$ & LPIPS$\downarrow$ & PSNR$\uparrow$ & SSIM$\uparrow$ & LPIPS$\downarrow$  \\
    \midrule
    NeRF-T \cite{li2021neural}  & 21.84 & 0.69 & 0.37 & 12.33  & 0.49 & 0.65 & 21.98 & 0.69 & 0.46 & 21.16 & 0.71 & 0.36 & 20.63 & 0.53 & 0.49 & 20.52 & 0.75 & 0.39 \\
    % \midrule
    HyperNeRF \cite{park2021hypernerf} & 16.43 & 0.43 & 0.40 & 16.81 & 0.41 & 0.56 & 16.04 & 0.38 & 0.62 & 17.64 & 0.42 & 0.43 & 18.52 &  0.39 & 0.52 & 16.75 & 0.51 & 0.23 \\
    
    D3DGS \cite{yang2023deformable3dgs} & 25.02 & \textbf{0.91} & 0.13 & 19.68 & 0.78 & 0.25 & 21.67 & 0.88 & 0.23 &  19.78 & 0.85 & 0.16 & 20.60 & 0.82 & 0.24 & 19.10 & 0.88 & 0.18 \\
    
    4DGS \cite{Wu_2024_CVPR} & 23.23 & 0.87 & 0.17 & 19.19 & 0.77 & 0.30 & 21.85 & 0.75 & 0.42 &  21.85 & 0.87 & 0.16 & 22.02 & 0.80 & 0.27 &  - & - & - \\
    
    \midrule
    
    Vid2Avatar \cite{guo2023vid2avatar} &  17.41 &  0.56 &  0.60 &  14.32 &  0.62 &  0.65 &  21.56 &  0.69 &  0.50 &  14.86 &  0.51 &  0.69 &  15.04 &  0.41 &  0.70 &  13.96 &  0.60 &  0.68 \\
    % \midrule
    NeuMan \cite{jiang2022neuman} & 23.99 & 0.78 & 0.26 & 24.63 & 0.81 & 0.26 & 25.43 & 0.80 & 0.31 &  25.55 & 0.83 & 0.23 & 22.70 & 0.68 & 0.32 & 24.96 & 0.86 & 0.21 \\
    
    % \midrule
    HUGS \cite{kocabas2024hugs} &  25.94 &  0.85 &  0.13 &  25.54 &  0.86 &  0.15 &  26.86 &  0.85 &  0.22 & 25.46 &  0.84 &  0.13 &  23.75 &  0.78 &  0.22 &  26.00 &  \textbf{0.92} &  0.09
    \\
    HUGS* \cite{kocabas2024hugs} &  26.26 &  0.85 &  0.10 &  25.68 &  0.86 &  \textbf{0.09} &  26.69 &  0.84 &  0.14 & 25.94 &  0.85 &  0.09 &  23.71 &  0.77 &  0.18 &  26.09 &  \textbf{0.92} &  \textbf{0.07}
    \\
    \midrule
    
    StM &  \textbf{27.60} &  \textbf{0.91} &  \textbf{0.06} &  \textbf{26.44} &  \textbf{0.87} &  \textbf{0.09} &  \textbf{27.49} & \textbf{0.88} &  \textbf{0.10} & \textbf{26.75} &  \textbf{0.88} &  \textbf{0.07} &  \textbf{24.57} &  \textbf{0.82} &  \textbf{0.15} &  \textbf{26.60} &  \textbf{0.92} &  \textbf{0.06}
    \\
    \bottomrule
    \end{tabular}
    }  
    % \vspace{-0.2cm}

    \label{tab:neuman_human_scene}
\end{table*}

\begin{table*}[htb!]
    \centering
    \caption{Comparing the novel view synthesis results of the \textbf{\em human} region on Neuman \cite{jiang2022neuman} dataset.
    % The human region is the crop based on human detection box.
    % HUGS* is our re-produced results trained for 20,000 iterations.
    % Performance is evaluated on PSNR, SSIM and LPIPS metrics.
    }
    \setlength{\tabcolsep}{3pt}
    \resizebox{\textwidth}{!}{
    \begin{tabular}{l|ccc|ccc|ccc|ccc|ccc|ccc}
    \toprule
        & \multicolumn{3}{c|}{\textbf{Seattle}} & \multicolumn{3}{c|}{\textbf{Citron}} & \multicolumn{3}{c|}{\textbf{Parking}} & \multicolumn{3}{c|}{\textbf{Bike}} & \multicolumn{3}{c|}{\textbf{Jogging}} & \multicolumn{3}{c}{\textbf{Lab}}   \\
    \midrule
        & PSNR $\uparrow$ & SSIM $\uparrow$ & LPIPS $\downarrow$ & PSNR $\uparrow$ & SSIM $\uparrow$ & LPIPS $\downarrow$ & PSNR $\uparrow$ & SSIM $\uparrow$ & LPIPS $\downarrow$ & PSNR $\uparrow$ & SSIM $\uparrow$ & LPIPS $\downarrow$ & PSNR $\uparrow$ & SSIM $\uparrow$ & LPIPS $\downarrow$ & PSNR $\uparrow$ & SSIM $\uparrow$ & LPIPS $\downarrow$ \\
    \midrule 
    Vid2Avatar \cite{guo2023vid2avatar} &  16.90 &  0.51 &  0.27 &  15.96 &  0.59 &  0.28 & 18.51 &  0.65 &  0.26 &  12.44 &  0.39 &  0.54 &  16.36 &  0.46 &  0.30 &  15.99 &  0.62 &  0.34 \\
    % \midrule
    NeuMan \cite{jiang2022neuman}    & 18.42 & 0.58 & 0.20 & 18.39 & 0.64 & 0.19 &  17.66 & 0.66 & 0.24 & 19.05 & 0.66 & 0.21 &  17.57 & 0.54 & 0.29 & 18.76 & 0.73 & 0.23 
    \\
    HUGS \cite{kocabas2024hugs}       &  19.06 &  0.67 &  0.15 &  19.16 &  \textbf{0.71} &  0.16 &  19.44 &  \textbf{0.73} &  0.17 &  19.48 &  0.67 &  0.18 & 17.45 &  0.59 &  0.27 &  18.79 &  0.76 &  0.18
    \\
    HUGS*  \cite{kocabas2024hugs}     &  18.80 &  0.64 &  0.13 &  18.99 &  0.70 &  0.13 &  19.14 &  0.72 &  0.15 &  19.82 &  0.67 &  \textbf{0.14} & 17.25 &  0.58 &  0.23 &  18.88 &  0.76 &  0.15
    \\
    \midrule
    StM       &  \textbf{19.93} &  \textbf{0.71} &  \textbf{0.12} &  \textbf{20.20} &  \textbf{0.71} &  \textbf{0.12} &  \textbf{19.80} &  \textbf{0.73} &  \textbf{0.14} &  \textbf{20.41} &  \textbf{0.70} &  \textbf{0.14} & \textbf{18.28} &  \textbf{0.62} &  \textbf{0.21} &  \textbf{19.73} &  \textbf{0.77} &  \textbf{0.14}
    \\
    \bottomrule
    \end{tabular}  
    }
    % \vspace{-0.2cm}

    \label{tab:neuman_human_crop}
\end{table*}

Given the presence of millions of Gaussian primitives, directly modeling cross-interactions between the scene and human avatar would be computationally prohibitive.
To address this challenge, we introduce a shared non-linear transformation function, specifically a residual MLP that maps each Gaussian primitive, whether belonging to the scene or the human avatar, into a unified integration space. This approach enables efficient and structured information exchange while maintaining computational feasibility.
Furthermore, each Gaussian primitive is associated with a set of attributes, including position, opacity, color, scale, and rotation. 
These attributes, each carrying distinct semantic significance, are freely optimized and require individualized processing.
Taking these factors into account, we extend transformation function to each Gaussian attributes, which enables a more structured and efficient projection between the background scene and the human avatar.

% These attributes can be freely optimized and represent own specific meaning, therefore they require individual processing. 

% Considering millions of Gaussian primitives, it will be resource exhausted if direct contain crossing interaction between the scene and human avatar.
% Therefore, we adopt shared residual MLP to mapping each Gaussian belong to the scene and human avatar to an integration space.

% Due to the separated modeling of scene and human avatar as above,
% the learning flexibility may cause the two models unaligned.
% To address this, we further introduce a model alignment process.

% Recall that, each Gaussian primitive comes with a series of features such as position, opacity, color, scale, and rotation. 
% Conceptually, all these features can be freely set without any correlation involved.
% That being said, these features should be processed individually.
% On the other hand, there is no correspondence or pairing relationship 
% at the Gaussian primitive level between scene and human avatar model.
% Putting all these factors together, we propose a simple yet effective {\em per-feature alignment} algorithm.

Specifically, for each attribute $i$ we introduce the shared transformation function $\phi_i$ (e.g., a shared residual MLP) as:
% At every optimizing iteration, we propose to add a residual MLP called alignment MLP for each property, including position, color, opacity, scale, and rotation, defined as: 
\begin{equation}
\label{eq:mlp}
% {G^{P}}' = G^{P} + Linear_{2}(ReLU(Linear_{1}(G^{P}))
% {G^{i}}' = G^{i} + \phi_i(G^{i}), \text{where}, G^{i} = Cat(G_{s}^{i}, G_{a}^{i})
% {G_{i}^{a}}' = G_{i}^{a} + \phi_i(G_{a}^{i}), \;\;
% {G_{i}^{s}}' = G_{i}^{s} + \phi_i(G_{s}^{i}),
{\mathcal{G}_{a}^{i}}' = \mathcal{G}_{a}^{i} + \phi_i(\mathcal{G}_{a}^{i}), \;\;
{\mathcal{G}_{b}^{i}}' = \mathcal{G}_{b}^{i} + \phi_i(\mathcal{G}_{b}^{i}),
\end{equation}
% where $G_{i}^{a}$ and $G_{i}^{s}$ denote the $i$-th Guassian feature 
% of the scene ($G_{s}$) and avatar ($G_{a}$) models. 
where $\mathcal{G}_{a}^{i}$ and $\mathcal{G}_{b}^{i}$ denote the $i$-th Guassian attribute of the background ($\mathcal{G}_{b}$) and avatar ($\mathcal{G}_{a}$) models. 
After being aligned,
the two models can be simply concatenated
and fed into the rendering process to produce the final images.

% $Cat$ specifies a concatenation function of Gaussian primitives
% for the scene ($G_{s}$) and avatar ($G_{a}$) models.

% \vspace{-9pt}
\subsection{Discussion}
\label{sec:discussion}
To demonstrate that our shared mapping module effectively exchanges information beyond a residual function, we compare our method with an alternative approach that applies separate residual MLPs for each Gaussian fields per-attribute.
Specifically, while our method uses a shared-weight MLP per attribute type that processes primitives from both background and avatar fields (enabling cross-model information flow through weight sharing),
the separate baseline uses independent MLPs with distinct weights for each field—one for background and one for avatar—resulting in a lack of information exchange between the two components.
As shown in Table~\ref{tab:ablation} (b) and (c), this separate approach leads to a significant performance drop compared to our shared method, validating that the shared-weight architecture is crucial for learning a unified transformation that properly aligns both sets of Gaussian primitives into a coherent representation space.

% Utilizing a shared MLP along the attribute dimension of each primitive, rather than across all primitives, has similar practice in PointNet \cite{qi2017pointnet, qi2017pointnetplusplus}. However, PointNet emphasizes the unordered nature of point clouds, our approach focuses on learning a nonlinear transformation that projects both sets of Gaussian primitives into a unified space.
Utilizing a shared-weight MLP along the attribute dimension of each primitive, rather than across all primitives, has similar practice in PointNet~\cite{qi2017pointnet, qi2017pointnetplusplus}. 
However, PointNet emphasizes the permutation-invariant processing of unordered point clouds through symmetric aggregation functions, whereas our approach focuses on learning a nonlinear transformation that projects both sets of separately initialized and optimized Gaussian primitives into a unified representation space. 
The key distinction is that our shared weights enable bidirectional information flow between foreground and avatar components during training: gradients from both fields jointly optimize the same transformation parameters, forcing the network to discover a common alignment strategy that works coherently for both components, thereby enabling effective integration of independently modeled human avatars and background scenes.

\subsection{Mapping network details}
\label{sec:implement_details}
We present a detailed overview of our shared information mapping module. Specifically, we adopts non-linear lightweight residual MLPs for each attribute, including position, SH, opacity, scale, rotation, defined as: 
\begin{equation}
\label{eq:mlp}
\phi_i = Linear_{2}(ReLU(Linear_{1}(\mathcal{G}^{i}))
\end{equation}
where $\mathcal{G}^{i}$ is the $i$-th attribute of Gaussian models. 
The input dimension corresponds to the dimension of each attribute, while the hidden dimension of $\phi_i$ is set to 64.

\subsection{Model training}
% We jointly optimize background Gaussians $\mathcal{G}_b$ and avatar Gaussians $\mathcal{G}_a$ with its feature triplane, and deformation modules, and also our inter-model alignment layers.

Our model is trained end to end, with the learnable parameters
distributed in the background scene $\mathcal{G}_b$ and avatar Gaussians $\mathcal{G}_a$ models, the feature triplane, deformation modules, as well as our information mapping module.
Specifically, we adopt standard image-based losses between rendered and ground truth images: $\mathcal{L}_{rbg}$ for RGB similarity, $\mathcal{L}_{ssim}$ based on the SSIM metric \cite{wang2004image}, and $\mathcal{L}_{lpips}$ for perceptual similarity \cite{zhang2018unreasonable}.
Following HUGS \cite{kocabas2024hugs}, we apply image-level losses ($\mathcal{L}^a_{rbg}$, $\mathcal{L}^a_{ssim}$$\mathcal{L}^a_{lpips}$) specifically to the avatar Gaussian rendering.

Additionally, each avatar Gaussian primitive's LBS weights are regularized by  $\mathcal{L}_{lbs}$ \cite{kocabas2024hugs, peng2021animatable}, ensuring they match the distance-weighted average of the $k$ nearest mesh vertices weights estimated by SMPL \cite{SMPL:2015}.
Specifically, we employ an $\ell_2$ loss to regularize the LBS weights $\bm{W}$ of the Gaussian points, ensuring they closely align with the LBS weights of nearby SMPL vertices.
For each individual Gaussian $\bm{p}_i$, we retrieve its $k=6$ nearest vertices on the SMPL mesh. A distance-weighted average of the LBS weights of these vertices is computed to obtain $\hat{\bm{W}}$. And the loss is 
\begin{equation}
    \mathcal{L}_{\text{LBS}} = \| \bm{W} - \hat{\bm{W}} \|_{\text{F}}^2.
\end{equation}

To further improve the global coherent representation,
we add a depth regularization loss $\mathcal{L}_{depth}$. 
To handle scale ambiguity in the monocular estimated depth map, we apply a \textit{pearson} correlation loss \cite{zhu2025fsgs}, that enforces only the linear relationship between the monocular estimated depth $\hat{D}_{\text{est}}$ and the rendered depth $\hat{D}_{\text{ras}}$ rather than their absolute values, as described by the following function:
\begin{equation}
    \mathcal{L}_{depth}=
    \operatorname{Corr}(\hat{D}_{\text{ras}}, \hat{D}_{\text{est}}) = \frac{\operatorname{Cov}(\hat{D}_{\text{ras}}, \hat{D}_{\text{est}})} {\sqrt{\operatorname{Var}(\hat{D}_{\text{ras}})\operatorname{Var}(\hat{D}_{\text{est}})}}
    \label{eq:depthloss}
\end{equation}
This relaxed constraint enables the alignment of depth structures without being obstructed by inconsistencies in absolute depth values.

As we demonstrated in Sec~\ref{sec:ablation}, simply increasing depth regularization is insufficient to solve our main proposed problem, as single-view depth estimation remains imprecise and noise, particularly in regions where the human subject interacts with the environment. 
% which only regularizes the linear relationship between monocular estimated depth and rendered depth rather than absolute value. 

The overall objective loss function is therefore:
\begin{equation}
\label{formula:loss}
\begin{split}
       \mathcal{L} = &\lambda_{rbg}\mathcal{L}_{rbg} + \lambda_{ssim}\mathcal{L}_{ssim} + \lambda_{lpips}\mathcal{L}_{lpips}\\
        + &\lambda_{rbg}\mathcal{L}_{rbg}^{a} + \lambda_{ssim}\mathcal{L}_{ssim}^{a} + \lambda_{lpips}\mathcal{L}_{lpips}^{a}\\
        + &\lambda_{lbs}\mathcal{L}_{lbs} + \lambda_{depth}\mathcal{L}_{depth},
\end{split}
\end{equation}
where we consistently set all hyperparameters as follows: $\lambda_{rbg}=0.8$, $\lambda_{ssim}=0.2$, $\lambda_{lpips}=1.0$, $\lambda_{lbs}=100$, $\lambda_{depth}=0.02$ across all the scenes.

\subsection{Novel view, novel pose and novel scene synthesis}

The background scene and human avatar models provide a rich environment for creating a diverse range of human-centric scenarios. Novel view synthesis for both the original scene and the human avatar is naturally supported. To compose a specific scenario, we begin by selecting a background scene—either the original or a new one—then set the motion sequence and relative relationship of the avatar within the scene. 
The human avatar is flexible animated by motion sequences with the help of SMPL \cite{SMPL:2015} model.
% The human avatar is animated by the poses with by the SMPL \cite{loper2023smpl} model, with flexible control over the pose trajectory. 
The final scenario is constructed by positioning the human avatar model according to its scene coordinates and integrating both models in their original form (Gaussian primitives). Standard rasterization-based rendering is then applied to generate a specific view of the whole scenario.

\section{Experiments}
\label{sec:experiment}
% \subsection{Dataset and Metrics}
\subsection{Experimental settings}

\noindent \textbf{Dataset}
We use {\em NeuMan} dataset~\cite{jiang2022neuman} with six video sequences, named Seattle, Citron, Parking, Bike, Jogging, and Lab, taken from in-the-wild environments using a moving mobile phone camera. 
% Each video contains a single person walking around and doing some natural movements, and it exhibits most human body regions.
We follow the official training and testing splits ~\cite{jiang2022neuman} to facilitate fair comparison with existing competitors.

\begin{figure*}
  \centering
   \includegraphics[width=1\linewidth]{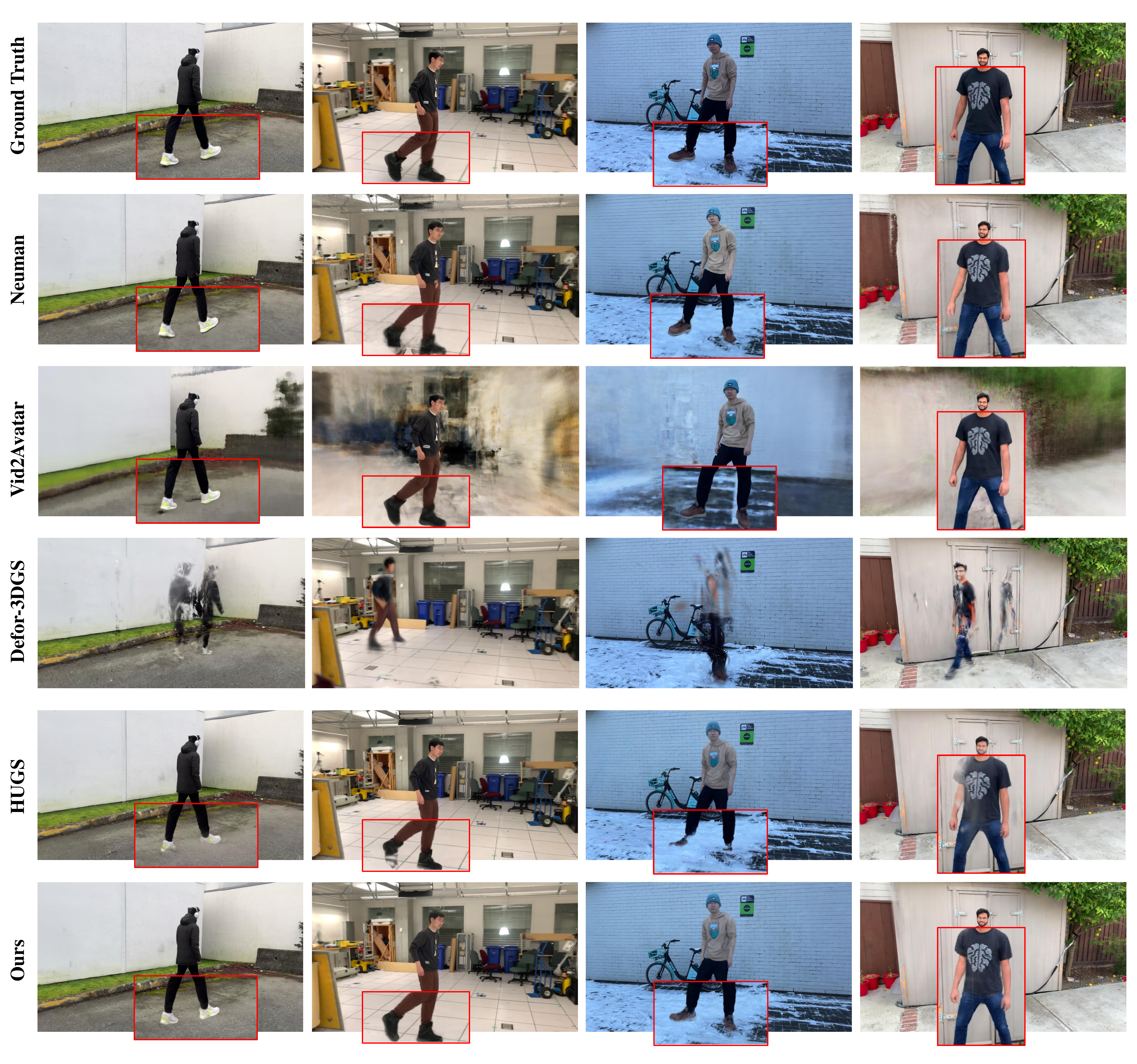}
   % \vspace{-0.8cm}
   \caption{Qualitative comparison for novel view synthesis comparing our StM with HUGS~\cite{kocabas2024hugs}, D3DGS~\cite{yang2023deformable3dgs}, Vid2Avatar~\cite{guo2023vid2avatar}, and Neuman~\cite{jiang2022neuman}. The zoomed-in regions (red box) highlight the difference.}
   \label{fig:neuman_comp_1}
   % \vspace{-0.3cm}
\end{figure*}

\noindent \textbf{Evaluation metrics.}
We consider three image quality metrics: including the peak signal-to-noise ratio (PSNR), structural similarity index measure (SSIM)~\cite{wang2004image}, and the learned perceptual image patch similarity (LPIPS)~\cite{zhang2018unreasonable}, which are broadly recognized standards in the field.

\noindent \textbf{Implementation details.} 
We employ the Adam optimizer~\cite{diederik2014adam} for training.
To optimize the position of Avatar Gaussians and Scene Gaussians,
we initialize the learning rate to $1.6^{-4}$ and terminate to $1.6^{-6}$, coupled with an exponential learning rate decay.
For the information mapping module,
we more heavily initialize the learning rate to $10^{-3}$ and terminate to $10^{-5}$, also using an exponential schedule.
% learning rate decay for the inter-model alignment module, and the same schedule from $1.6^{-4}$ to $1.6^{-6}$ to optimize the Avatar Gaussians and Scene Gaussians position.
During the optimization process, we clone, split, and prune scene Gaussian field every 100 iterations~\cite{kerbl3Dgaussians}, and these operations are applied to the avatar Gaussian fields every 600 iterations~\cite{kocabas2024hugs}.
% and reset the scene Gaussians opacity in front of avatar Gaussian kernel every 500 iterations from 3000 to 15000~\cite{yang2024localized}.
The entire optimization process runs for 20,000 iterations on a single NVIDIA RTX 3090 GPU.
% , with training times for each scene ranging from 0.5 to 2 hours.
All presented results for GS-based methods are averaged over three runs with different seeds.

\begin{figure*}[!b]
  \centering
   \includegraphics[width=1\linewidth]{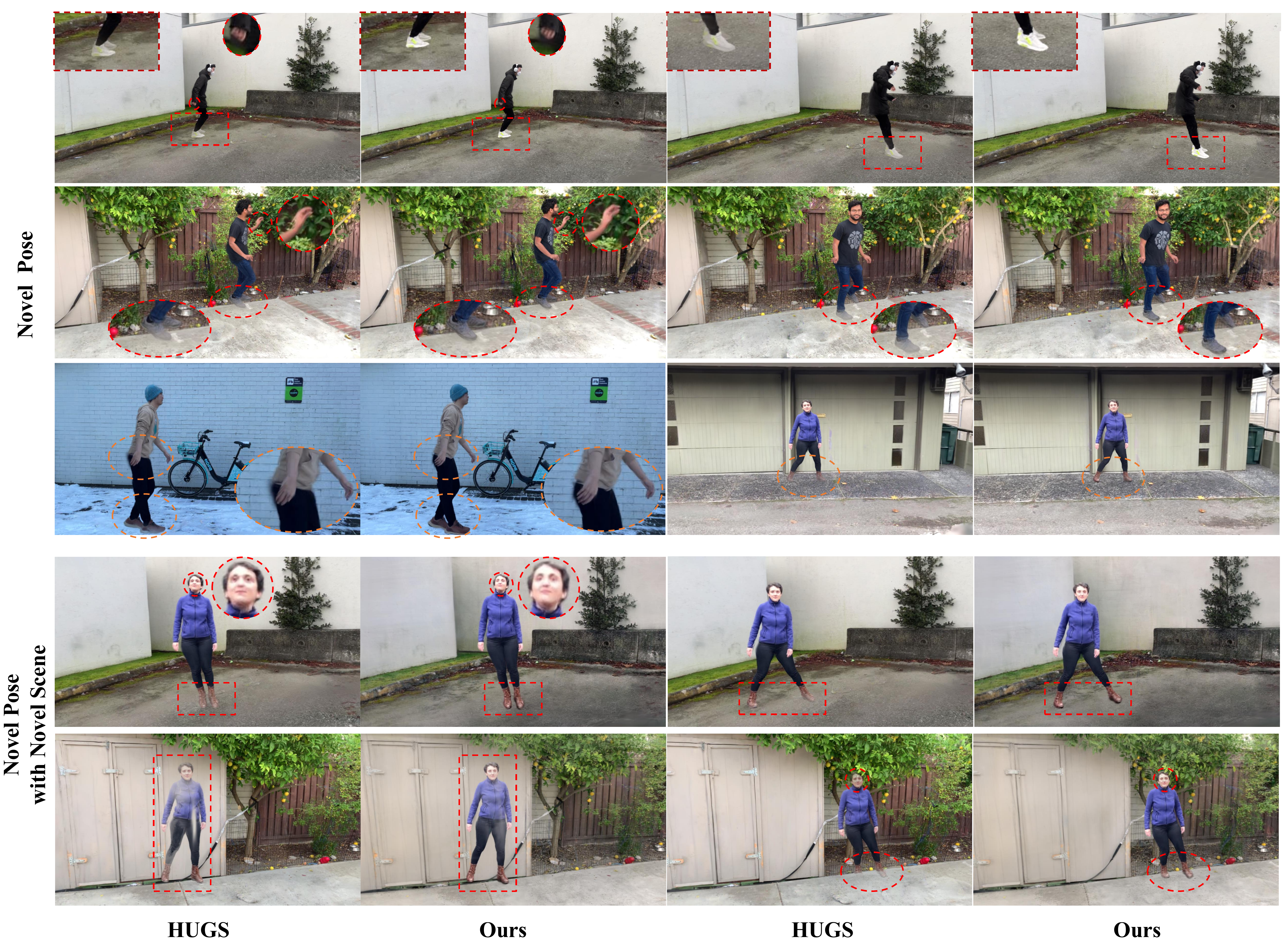}
   % fig4-neumananim.pdf
   % \vspace{-0.8cm}
   \caption{Qualitative evaluation for novel pose synthesis and novel scene composition. Our method generates high-quality results that maintain the fine details and integrity of the avatar, and showing reduced floating artifacts in the background scene.}
   \label{fig:neuman_novelpose}
\end{figure*}

\subsection{Quantitative analysis}

% \noindent \textbf{Competitors.}
\noindent \textbf {Novel view synthesis evaluation.}
We compare the novel view synthesis results by our method and existing state of the art alternatives in the categories of
% including
NeRF-based whole dynamic scene reconstruction methods: NeRF-T~\cite{li2021neural}, HyperNeRF~\cite{park2021hypernerf}; 
NeRF-based detached modeling methods: Neuman~\cite{jiang2022neuman}, Vid2Avatar~\cite{guo2023vid2avatar}; 
Gaussian-based whole dynamic scene reconstruction methods: Deformable 3D Gaussian (D3DGS)~\cite{yang2023deformable3dgs}, 4DGS~\cite{Wu_2024_CVPR}; and Gaussian-based detached method: HUGS~\cite{kocabas2024hugs}.

The novel view synthesis results for the whole scenes are presented in Table~\ref{tab:neuman_human_scene}.
We highlight several key observations:
(1) By modeling the whole scene using a single representation model,
previous methods all face challenges for novel view rendering, regardless using either NeRF (NeRF-T~\cite{li2021neural}, HyperNeRF~\cite{park2021hypernerf} or Gaussian splatting (4DGS~\cite{Wu_2024_CVPR}, D3DGS~\cite{yang2023deformable3dgs}).
In particular, HyperNeRF is the worst of dealing with drastically dynamic scenes with humans, likely due to its reliance of dense multi-view images.
(2) Although a couple of methods (Neuman~\cite{jiang2022neuman} and Vid2Avatar~\cite{guo2023vid2avatar}) learn the background and human avatar separately, a
limited performance gain is achieved.
As they suffer from inferior background optimization (Vid2Avatar~\cite{guo2023vid2avatar})
or the challenge of modeling photorealistic the human body parts (Neuman~\cite{jiang2022neuman}).
% (3) Reasonably addressing these issues, HUGS~\cite{kocabas2024hugs} clearly improve the results of novel view rendering, mainly due to the introduction of human parametric model aware 3DGS based representation model that can be optimized end to end.
(3) By effectively addressing these challenges, HUGS~\cite{kocabas2024hugs} significantly enhances the quality of novel view rendering. This improvement is primarily attributed to the introduction of the human parametric model-aware 3DGS-based representation, which enables end-to-end optimization.
However, this approach overlooks the integration issue between the separately modeled background and foreground.
(4) By tackling this problem and increasing the shared information mapping module,
our StM strategy further enhances the rendering performance substantially.

% achieves more fluid human avatar models along with good quality of background scene,
% thanks to its end to end learnable pose aware representation.
% However, they still suffer from several issues, such as rendering mixture of background and human from region to region, the floating effect, blurring boundary and artifacts around the human avatar.
% This indicates that the foreground and background models 
% are not aligned for the rendering process,
% as they are parameterized individually and optimized 
% in parallel without effective interaction.

\noindent \textbf {Human-scene interaction region evaluation.}
We further focus on the quality of the human-scene interaction region by cropping the human using a human detector~\cite{Guler2018DensePose} from the entire testing images. 
This evaluation specifically targets the challenging boundary areas where avatar and background meet—regions most prone to floating artifacts, occlusion errors, and blending inconsistencies.
As shown in Table~\ref{tab:neuman_human_crop}, 
we observe that StM also achieves a significant performance improvement over the competitors, indicating that our shared information mapping mechanism successfully mitigates boundary artifacts and ensures spatial coherence in the critical interaction regions.

% we evaluate only human regions on the test view, which is the tight crop around human.
% Our method improvement on the human-crop region is even larger than the whole scene, indicating that our design improves the interaction regions of the human with the background more.

% In Table~\ref{tab:neuman_human_scene}, we evaluate the rendering quality in test views.
% NeRF-T~\cite{li2021neural}, HyperNeRF~\cite{park2021hypernerf}, 4D-GS~\cite{Wu_2024_CVPR}, Deformable 3D Gaussian~\cite{yang2023deformable3dgs} are not specifically designed for the human avatar, will show poor reconstruction quality.
% Neuman~\cite{jiang2022neuman} and Vid2Avatar~\cite{guo2023vid2avatar} both take a long time to train, they show improved results, but still not as good as Gaussian Splatting methods~\cite{kocabas2024hugs}.
% Overall, our method gives the best rendering quality.

% In Table~\ref{tab:neuman_human_crop}, we evaluate only human regions on the test view, which is the tight crop around human.
% Our method improvement on the human-crop region is even larger than the whole scene, indicating that our design improves the interaction regions of the human with the background more.

\noindent \textbf {Comparison with avatar-only reconstruction methods.}
To further validate our approach, we compare against recent avatar-only reconstruction 
methods that focus solely on human modeling without explicit scene integration: GaussianAvatar~\cite{hu2024gaussianavatar} and GauHuman~\cite{hu2024gauhuman}. 
Unlike the interaction region evaluation above (which crops humans from full scene reconstructions), this comparison evaluates methods that only model avatars, typically using foreground masks to isolate humans during training.
As shown in Table~\ref{tab:rebuttal_label}, even when evaluating only the avatar, our method substantially outperforms these avatar-specialized approaches.
This significant gap demonstrates that modeling the human and scene jointly—rather than in isolation—provides stronger constraints and better handles the challenging aspects of monocular reconstruction such as limited viewpoints and complex human-scene interactions. These results validate our motivation that holistic scene modeling with proper integration mechanisms is essential for high-quality human-centric reconstruction.

% \begin{table}
%     \centering
%     \begin{tabular}{cccc}
%          Bike & PSNR $\uparrow$ & SSIM $\uparrow$ & LPIPS $\downarrow$ \\
%          GaussainAvatar & 7.95 & 0.47 & 0.45\\
%          Ours & 20.41 & 0.70 & 0.14 \\
%          &  &  & \\
%     \end{tabular}
%     \caption{Caption}
%     \label{tab:my_label}
% \end{table}

\begin{table}
    \centering
    \caption{Avatar-only comparison on Neuman Dataset}
    \begin{tabular}{l|ccc}
        \toprule
        Model    & PSNR $\uparrow$ & SSIM $\uparrow$ & LPIPS $\downarrow$ \\
        \midrule
        GaussainAvatar & 25.10 & 0.96 & 0.040 \\
        GauHuman & 22.12 & 0.93 & 0.046\\
        Ours & \textbf{30.42} & \textbf{0.97} & \textbf{0.022} \\
        \bottomrule
    \end{tabular}
    \label{tab:rebuttal_label}
\end{table}

Moreover, our StM strategy only increases MLP layers with a limited number of parameters, therefore, the additional computational time introduced is minimal. Specifically, it increases by only 0.001 seconds per iteration, resulting in an overall increase of 20 seconds per scene. 
This demonstrates that our method achieves significant quality improvements with negligible additional cost.

\subsection{Qualitative analysis}
\begin{figure*}
  \centering
   \includegraphics[width=1\linewidth]{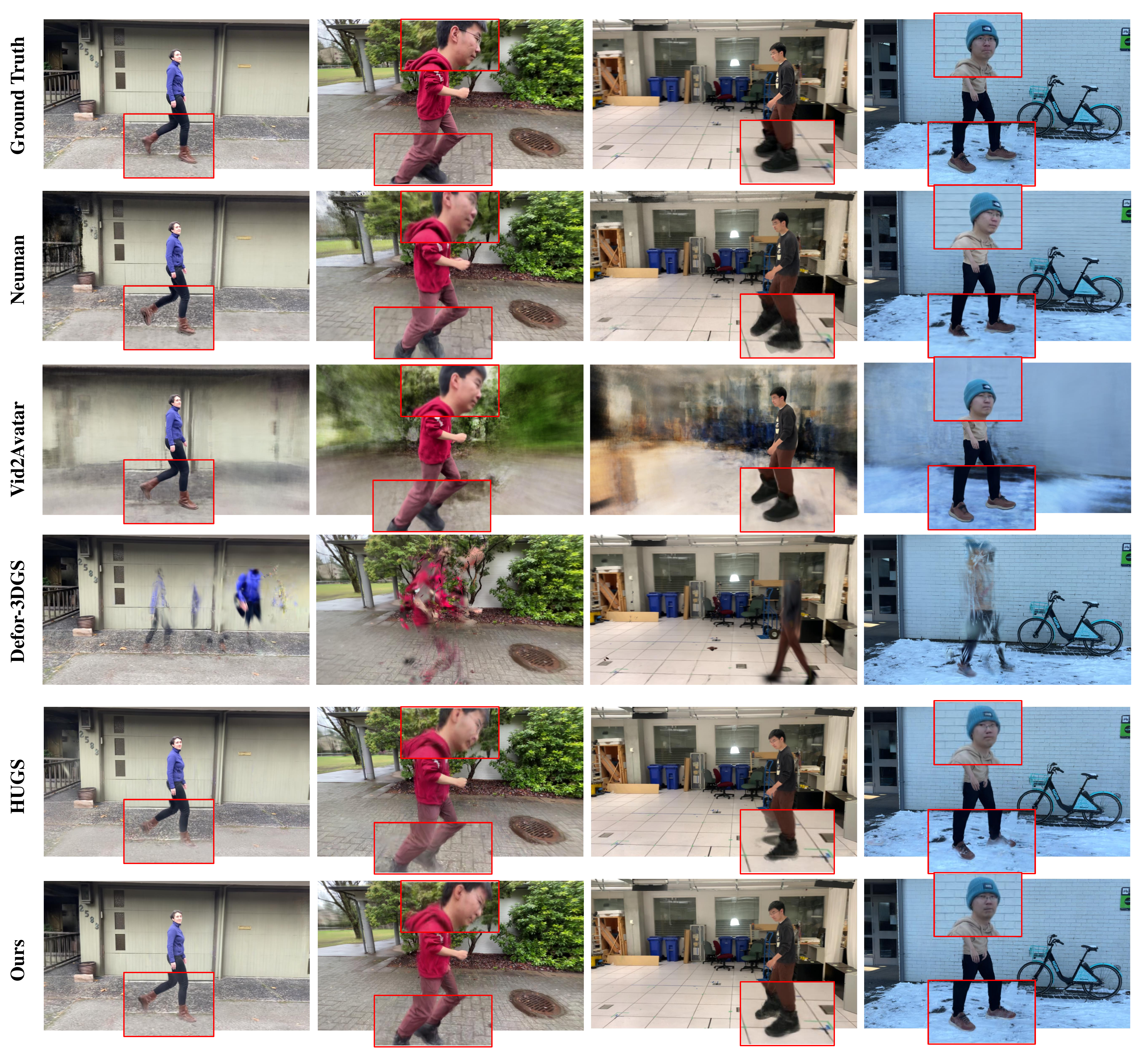}
   \caption{More qualitative comparison for novel view synthesis comparing our StM with HUGS~\cite{kocabas2024hugs}, D3DGS~\cite{yang2023deformable3dgs}, Vid2Avatar~\cite{guo2023vid2avatar}, and Neuman~\cite{jiang2022neuman}. The zoomed-in regions (red box) highlight the difference.}
   \label{fig:neuman_comp_2}
\end{figure*}

\noindent \textbf{Novel view synthesis evaluation.} 
In Figure~\ref{fig:neuman_comp_1} and Figure~\ref{fig:neuman_comp_2},
we show the qualitative results compared with Neuman~\cite{jiang2022neuman}, Vid2Avatar~\cite{guo2023vid2avatar}, D3DGS~\cite{yang2023deformable3dgs}, HUGS~\cite{kocabas2024hugs}.
We make the following observations:
(1) Without treating the background scene and foreground human avatar separately, D3DGS~\cite{yang2023deformable3dgs} even fails to model the intact human body.
% , leading to ghost effects and incomplete human instances.
This is plausibly due to not utilizing parametric human prior.
% the mixture of background and foreground models.
(2) By separately modeling the human from the background scene,
Vid2Avatar~\cite{guo2023vid2avatar} can achieve whole body rendering, but degrading the quality of the background, due to their foreground focused modeling design.
% suffer a huge performance degradation due to not have direct information to keep Gaussian points which represent avatars moving consistently and holistically, therefore the body will display divided in novel views.
(3) While Neuman~\cite{jiang2022neuman} improves background modeling, it still faces issues like stiff and unnatural effects.
A plausible reason is due to the intrinsic difficulty of implicit representation model (NeRF) in dealing with structural objects like human.
% get better reconstruction quality, but rendered images look very stiff and unnatural.
(4) To leverage explicit representation model like HUGS~\cite{kocabas2024hugs} achieves more fluid human avatar models along with good quality of background scene,
thanks to its combination with learnable parametric mesh representation.
However, they still doesn't work well especially where avatar interact with the background, such as rendering mixture of background and human from region to region, the floating effect, blurring boundary and artifacts around the human avatar.
This indicates that the foreground and background models are not belong to same representation space during the rendering process,
as they are parameterized individually and optimized with different strategies without effective interaction.
(5) By introducing the shared information mapping for per-attribute appropriately,
our StM strategy addresses this issue effectively and efficiently, as demonstrated by significantly improved novel view synthesis with finer details, reduced ghost effects, more complete human body, and better photo realism.

% Vid2Avatar~\cite{guo2023vid2avatar} shows blurry background rendering and Neuman~\cite{jiang2022neuman} get better reconstruction quality, but rendered images look very stiff and unnatural.
% And HUGS~\cite{kocabas2024hugs} achieve more fluid rendering for both the background and avatar, however, they exhibit occlusion of the human around the interaction regions. This mainly due to when the foreground and background Gaussian Splatting models are optimized simultaneously via gradient descent only following pixel level loss, local minima are likely to arise due to limited observations from the video. This leads to many floaters in the scene, which cannot be solved by continuing the optimization. In addition, as two models Gaussian points in the interactive regions have conflicts with each other, which can lead to undesirable artifacts and effects.
% In comparison, our method demonstrates better reconstruction quality and preserves fine details, as shown in the zoomed-in regions, especially improve foot interaction with the ground and addresses occlusion problem, and also improve the color consistence with environment lighting.
% This all achieved by our {\bf StA} training strategy, which can optimize Gaussians representation and further support optimizing.

% \begin{figure}[t]
%   \centering
%   % \fbox{\rule{0pt}{2in} \rule{0.9\linewidth}{0pt}}
%    \includegraphics[width=1\linewidth]{imgs/suppl2.pdf}
%    \caption{More qualitative rendering for novel pose synthesis comparing our StM with HUGS~\cite{kocabas2024hugs}.}
%    \label{fig:neuman_pose_supp}
% \end{figure}

\begin{figure}
  \centering
   \includegraphics[width=1\linewidth]{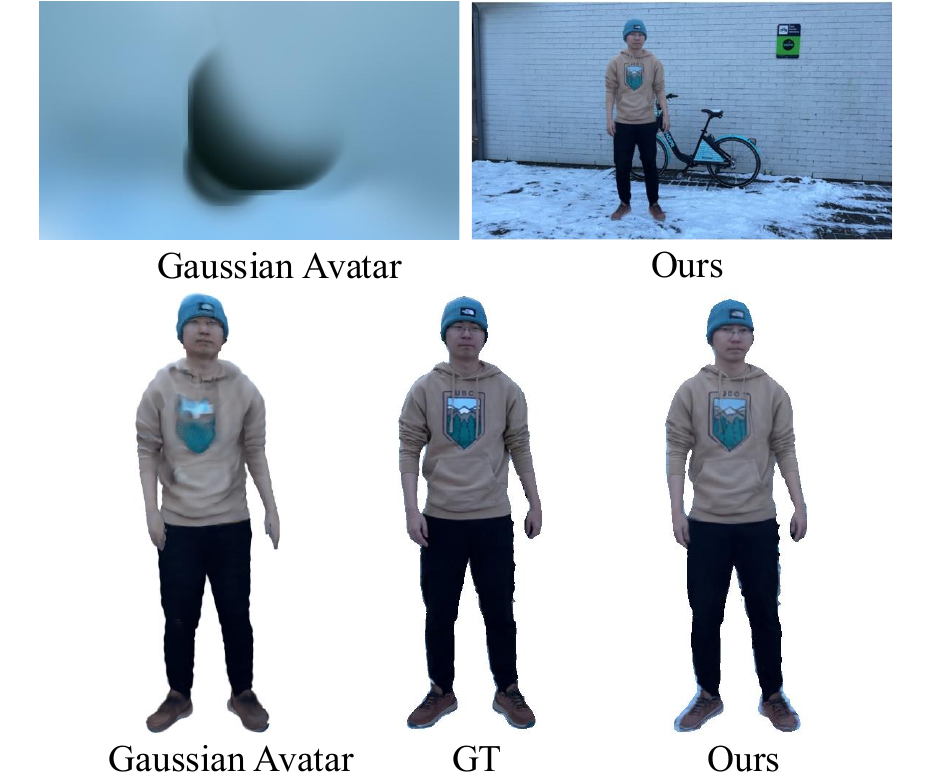}
   \caption{GaussianAvatar cannot reconstruct the entire scene, also rendering inferior avatar.}
   \label{fig:avatar-only-comparison}
\end{figure}

\noindent \textbf{Comparison with avatar-only reconstruction methods}. 
To further validate our holistic scene modeling approach, we compare against recent avatar-only reconstruction methods that focus solely on human modeling: GaussianAvatar~\cite{hu2024gaussianavatar}. As shown in Figure~\ref{fig:avatar-only-comparison}, GaussianAvatar 
cannot reconstruct the entire scene and produces inferior avatar quality compared to our method.
This validates our core motivation that holistic scene modeling with proper integration mechanisms is essential for high-quality human-centric reconstruction.

\noindent \textbf{Novel poses and novel scene evaluation.}
% We present a comparison of novel pose synthesis between StM and HUGS~\cite{kocabas2024hugs}
% As shown in Figure~\ref{fig:neuman_novelpose}, 
% HUGS often exhibits undesired foggy effects.
% In contrast, our StM renders the human avatar more complete.
% Importantly, this advantage can extend to the more challenging case with the human avatar transferred further to novel scene.
We present a comparison of novel pose synthesis and novel scene composition between StM and HUGS~\cite{kocabas2024hugs}. As shown in Figure~\ref{fig:neuman_novelpose}, HUGS exhibits several limitations: undesired foggy and ghosting effects in the rendered avatars (particularly visible in the torso and limbs), incomplete reconstruction of body parts such as legs and feet (highlighted in red boxes), and background contamination with floating artifacts near avatar-scene boundaries. In contrast, our StM renders the human avatar with significantly improved completeness, sharper boundaries, and cleaner separation from the background. 
Importantly, these advantages extend to the more challenging case of novel scene composition (bottom two rows), where avatars are transferred to entirely new environments. While HUGS struggles with increased blurriness and incomplete rendering in novel scenes, our method maintains avatar fidelity and generates more photorealistic compositions. This demonstrates that our shared information mapping mechanism learns more robust and generalizable avatar representations that reliably compose with novel backgrounds, validating the effectiveness of our approach across the full spectrum of human-centric 4D reconstruction tasks.

\begin{figure*}[!b]
  \centering
   \includegraphics[width=1\linewidth]{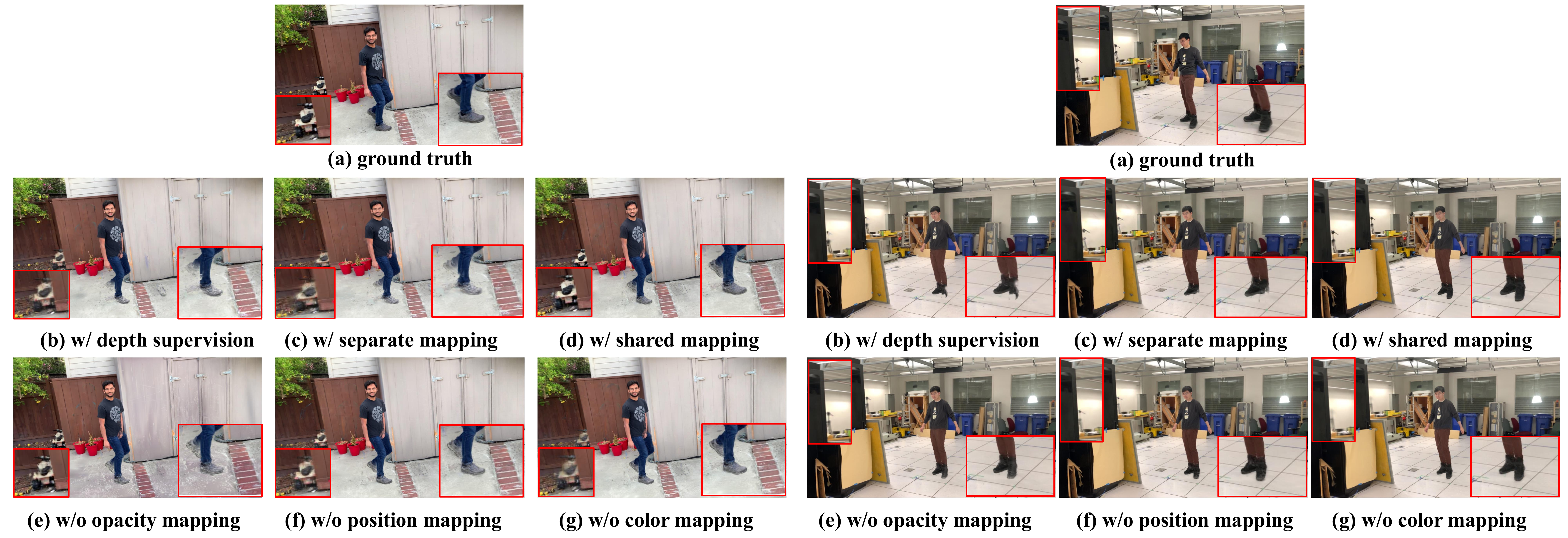}
   \caption{Qualitative ablation study on mapping strategies and per-attribute alignment. (a) Ground truth. (b) Depth supervision improves over baseline but exhibits blur at boundaries. (c) Separate mapping per field produces semi-transparent regions and blurred boundaries. (d) Our shared-weight mapping achieves clean, well-defined integration. (e)-(g) Ablations removing individual attribute mappings: without opacity mapping (e), severe ghosting appears; without position mapping (f), floating artifacts emerge; without color mapping (g), slight boundary blur occurs. Red boxes highlight critical interaction regions.}
   % \caption{Qualitative ablation study of alignment strategies and different Gaussian Splatting attribute alignments.}
   \label{fig:ablation}
\end{figure*}

\begin{figure*}
  \centering
   \includegraphics[width=0.8\linewidth]{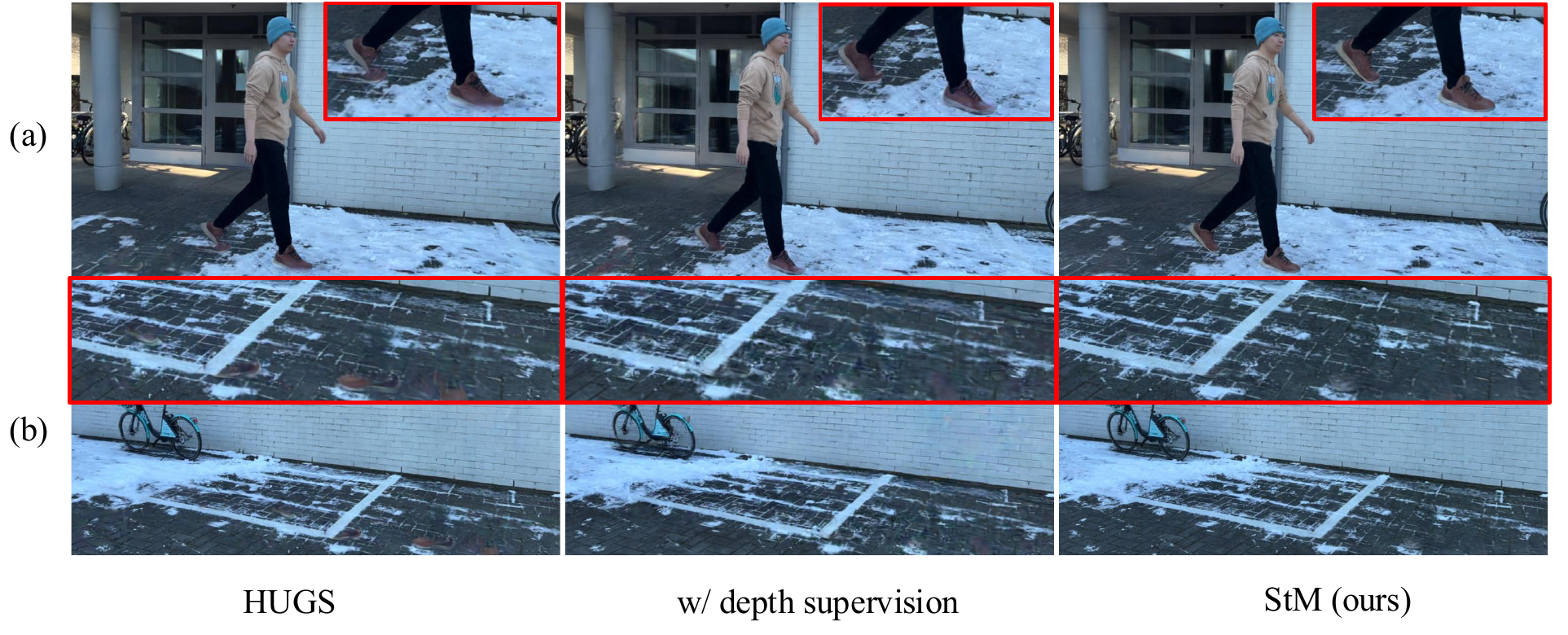}
    \caption{Depth regularization analysis. (a) Full scene rendering: HUGS exhibits severe blur at human-scene boundaries (feet/ground contact, highlighted in red boxes). Depth supervision improves structure but still produces blurred boundaries compared to our method. (b) Background-only rendering: HUGS with depth supervision leaves visible avatar residue (ghostly imprints in red boxes) in the background, revealing that depth regularization cannot ensure clean foreground-background separation. Our method achieves sharp boundaries and artifact-free backgrounds.}
   
   % \caption{
   % Further ablation.
   % % results demonstrate the effectiveness of our proposed module.
   % %Ablation on depth-only and ours shows that our method reduces avatar occlusion (a) and eliminates background artifacts %
   % (a) HUGS exhibits severe blur effect, depth helps but still clearly falls behind ours. 
   % %, w/ depth supervision, and w/ separate mapping, shows severe occlusion in interaction regions.
   % %(b)—caused by static scenes mistakenly retaining parts of dynamic objects (For better display, we only render the background). 
   % (b) HUGS could leave residue of avatar to the background. 
   % % (c) GaussianAvatar cannot reconstruct the entire scene, 
   % % also rendering inferior avatar.
   % % and even focusing only on the avatar will produce low-fidelity results.  
   % }
   \label{fig:rebuttal-1}
\end{figure*}

\begin{table*}[!t]
    \centering
    \caption{%Analysis of the effects of different choices 
    Ablation study of (a) Impact of estimated depth regularization; (b) Impact of our proposed shared information mapping module (without depth); (c) Results for separate MLPs instead of shared MLPs; (d)-(f) Effects of per Gaussian attribute mapping; (g) Full model (shared mapping with depth regularization). }
    \setlength{\tabcolsep}{3pt}
    \resizebox{\textwidth}{!}{
    \begin{tabular}{l|ccc|ccc|ccc|ccc|ccc|ccc}
    \toprule
        & \multicolumn{3}{c|}{\textbf{Seattle}} & \multicolumn{3}{c|}{\textbf{Citron}} & \multicolumn{3}{c|}{\textbf{Parking}} & \multicolumn{3}{c|}{\textbf{Bike}} & \multicolumn{3}{c|}{\textbf{Jogging}} & \multicolumn{3}{c}{\textbf{Lab}}   \\
    \midrule
        & PSNR $\uparrow$ & SSIM $\uparrow$ & LPIPS $\downarrow$ & PSNR $\uparrow$ & SSIM $\uparrow$ & LPIPS $\downarrow$ & PSNR $\uparrow$ & SSIM $\uparrow$ & LPIPS $\downarrow$ & PSNR $\uparrow$ & SSIM $\uparrow$ & LPIPS $\downarrow$ & PSNR $\uparrow$ & SSIM $\uparrow$ & LPIPS $\downarrow$ & PSNR $\uparrow$ & SSIM $\uparrow$ & LPIPS $\downarrow$ \\
    \midrule 

    (a) w/ depth & 26.39 & 0.8504 & 0.0964 & 26.06 & 0.8634 & 0.0892 & 26.99 &  0.8447 & 0.1373 & 26.10 & 0.8527  & 0.0941 & 23.75 & 0.7677 & 0.1861 &  26.17 & 0.9171 &  0.0689 \\

    (b) w/ shared mapping & 27.44 & 0.9068 & 0.0650 & 26.26 & 0.8661 & 0.0879 & 27.21 & 0.8766 & 0.1087 & 26.38 & 0.8799  & 0.0767 & 24.51 & 0.8206 & 0.1521 &  26.51 & 0.9215  &  0.0648 \\
    
    (c) separate MLPs & 26.63  & 0.8721  &  0.0794 & 25.85  & 0.8585 & 0.0926  &  26.12 & 0.8586  & 0.1252  &  25.63 & 0.8374 & 0.1084 & 24.00  & 0.7988 & 0.1706 &  25.24  &  0.8863  & 0.1181  \\

    \midrule
    
    (d) w/o position & 27.58 &  0.9096 & 0.0633  & 26.37  & 0.8589 & 0.0883 & 27.24  & 0.8643 & 0.1200 & 26.58 & 0.8819 & 0.0737 &  24.34 & 0.8137 & 0.1571 & 26.44 & 0.9149  &  0.0734
    \\
    
    (e) w/o opacity & 27.01  & 0.8909  & 0.0734 & 24.43  & 0.8499 & 0.1058 & 26.95 & 0.8601 & 0.1195 & 26.20 & 0.8781 & 0.0750 & 24.05 & 0.7995 & 0.1574& 26.31 & 0.9143 & 0.0720
    \\
   
    (f) w/o color & 27.58  & 0.9094  & 0.0655 &  26.24 & 0.8581 & 0.0881 & 27.20 & 0.8623 & 0.1246 & 26.41 & 0.8789 & 0.0765 &  24.30  &  0.8091  & 0.1521 &  26.39  & 0.9145  &  0.0720
    \\
    \midrule

    (g) Full &  \textbf{27.60} &  \textbf{0.9097} &  \textbf{0.0620} &  \textbf{26.44} &  \textbf{0.8674} &  \textbf{0.0874} &  \textbf{27.49} & \textbf{0.8771} &  \textbf{0.1049} & \textbf{26.75} &  \textbf{0.8836} &  \textbf{0.0705} &  \textbf{24.57} &  \textbf{0.8226} &  \textbf{0.1510} &  \textbf{26.60} &  \textbf{0.9225} &  \textbf{0.0634}

    % (d) Full &  \textbf{27.70} &  \textbf{0.9118} &  \textbf{0.0612} &  \textbf{26.43} &  \textbf{0.8698} &  \textbf{0.0817} &  \textbf{27.40} & \textbf{0.8672} &  \textbf{0.1184} & \textbf{26.70} &  \textbf{0.8828} &  \textbf{0.0734} &  \textbf{24.41} &  \textbf{0.8133} &  \textbf{0.1566} &  \textbf{26.44} &  \textbf{0.9159} &  \textbf{0.0720}
    \\
    \bottomrule
    \end{tabular}  
    }

    \label{tab:ablation}
\end{table*}

% We show the effect of ablating over our method in Table~\ref{tab:ablation} and Figure~\ref{fig:ablation}.
% We conduct ablation studies focus on the model alignment of human avatar and background scene.
% with qualitative results presented in Table~\ref{tab:ablation}
% and quantitative results in Figure~\ref{fig:ablation}.

\subsection{Ablation study}
\label{sec:ablation}

We conduct comprehensive ablation experiments to validate the effectiveness of each component in our method. 
All experiments are performed on the NeuMan dataset following the same protocol as our main evaluation.
Table~\ref{tab:ablation} presents quantitative results: (a) evaluates depth regularization alone, (b) evaluates our shared mapping mechanism without depth, (c) compares against separate MLPs baseline, (d)-(f) ablate individual attribute mappings, and (g) shows our complete model with both shared mapping and depth regularization.

\noindent \textbf{Effect of depth regularization.}
Depth regularization is a common approach to improving geometric consistency. 
As shown in Table~\ref{tab:ablation}(a), incorporating monocular depth supervision does provide measurable benefits over the baseline HUGS.
However, qualitative analysis 
reveals critical limitations.
Figure~\ref{fig:ablation}(b) shows that depth supervision still produces semi-transparent regions and spatial inconsistencies at ground contacts, with residual avatar traces contaminating background regions. 
This occurs because monocular depth estimation suffers from scale ambiguity and noise, particularly in dynamic regions and occlusion boundaries.
Figure~\ref{fig:rebuttal-1} provides more comprehensive evidence through 
two views.
In (a), full scene rendering shows HUGS with depth supervision still exhibits severe blur around the feet and lower body. 
The zoomed insets reveal the feet appear to "sink into" or "float above" the ground, 
indicating geometric constraints alone cannot resolve spatial inconsistencies without explicit model-to-model alignment. 
More revealing is (b), showing background-only rendering: HUGS with depth supervision clearly leaves visible residue artifacts—ghostly imprints of legs and feet persist in clean background regions (red box). 
This reveals depth regularization encourages geometric consistency within each model but does nothing to ensure clean separation between separately modeled avatar and background fields.
The monocular depth estimator cannot distinguish which primitives belong to foreground versus background in interaction regions, causing ambiguous assignments and contamination.

In contrast, our shared mapping explicitly addresses model-to-model correspondence by projecting both sets of primitives into unified space. 
As evident in both views, our method produces clean boundaries, eliminates residue artifacts, and maintains proper separation.
These findings demonstrate that while depth provides weak geometric priors, it cannot substitute for explicit information exchange mechanisms. The noise and uncertainty in monocular depth estimation—especially in challenging in-the-wild captures—make it unreliable for interaction regions where integration quality matters most.
Our alignment strategy, operating directly on Gaussian representations through learned transformations, proves far more effective.

\noindent \textbf{Effect of mapping strategy.}
To verify the effectiveness of our shared-weight mapping design, we compare against an alternative that applies separate residual MLPs for each Gaussian field per-attribute.
To verify the effectiveness of our shared-weight mapping design, we compare our method (Table~\ref{tab:ablation}(b)) against an alternative that applies separate residual MLPs for each Gaussian field per-attribute (Table~\ref{tab:ablation}(c)). 
Our shared mapping substantially outperforms the separate way across all scenes (e.g.,27.44 vs. 26.63 PSNR on Seattle, 26.26 vs. 25.85 on Citron).
Figure~\ref{fig:ablation}(c) and (d) illustrate this difference qualitatively: the separate strategy (c) produces semi-transparent regions and blurred boundaries at the avatar's feet and lower body (red boxes), while our shared mapping (d) achieves clean, well-defined boundaries. 
This validates that shared-weight architecture is essential for coherent integration of separately modeled components. Further discussion on the information exchange mechanism is provided in Section~\ref{sec:discussion}.

\noindent \textbf{Effect of individual per-attribute mapping.}
We evaluate the necessity and contribution of each individual attributes mapping design.
% This validates the contribution of each individual attributes in mapping.
% one specific alignment network for each attribute.
Here we examine three important ones: position, opacity, and color. 
Table~\ref{tab:ablation}(d)-(f) present ablations where we remove mapping for position, opacity, and color respectively.
We observe that opacity mapping is the most critical component among all attributes, as expected since it directly controls visibility and the blending of overlapping Gaussians.
As shown in Figure~\ref{fig:ablation}(e), without opacity alignment, severe ghosting effects and lighting inconsistencies appear at human-scene boundaries.
The absence of color alignment (Figure~\ref{fig:ablation}(g)) results in slight background blurring, while missing position alignment (Figure~\ref{fig:ablation}(f)) introduces inconspicuous floating artifacts and occlusion in boundary regions. These results confirm that each attribute requires individualized alignment processing to achieve optimal integration quality.

\section{Limitations}
\label{sec:limit}

While our method demonstrates significant improvements in dynamic human-centric scene reconstruction, several challenges in monocular video reconstruction remain. 
First, our approach is sensitive to upstream estimations: inaccurate camera poses and imperfect SMPL fitting can propagate errors, causing misalignments and artifacts in complex interaction regions. 
Second, monocular setups suffer from limited observability—back-side views are severely under-constrained, resulting in incomplete geometry and texture hallucination in unobserved regions. 
Despite these limitations, our extensive evaluation demonstrates that StM advances the state-of-the-art in this challenging domain. 
Future work could explore multi-view priors, foundation models for robust pose and depth estimation, or stronger geometric constraints to further improve reconstruction quality.

\section{Conclusion}
\label{sec:con}
We introduce {\bf StM}, a novel end-to-end strategy for dynamic human-centric scene  reconstruction from monocular videos.
Unlike prior approaches that either model scenes holistically or treat foreground and background as independent components with simple concatenation, we identify and address a critical overlooked challenge: the lack of effective information exchange between separately modeled human avatars and background scenes. 
Our key contribution is a shared information mapping mechanism that projects separately designed components into a unified representation space, enabling coherent integration while maintaining computational efficiency. 
By applying lightweight residual MLPs to each Gaussian attribute, our approach achieves optimal scene modeling without exhaustive pairwise interactions between millions of primitives. 
Extensive experiments demonstrate that our method significantly outperforms state-of-the-art alternatives across novel view synthesis, novel pose synthesis, and novel scene composition. 
Qualitative results validate that StM effectively mitigates floating artifacts, occlusion issues, and boundary inconsistencies at challenging human-scene interaction regions.
Our work establishes a foundation for integrated human-environment modeling with applications in AR/VR and digital human animation.

\ifCLASSOPTIONcaptionsoff
  \newpage
\fi

\bibliographystyle{IEEEtran}
\bibliography{main}

@article{qi2017pointnetplusplus,
  title={PointNet++: Deep Hierarchical Feature Learning on Point Sets in a Metric Space},
  author={Qi, Charles R and Yi, Li and Su, Hao and Guibas, Leonidas J},
  journal={arXiv preprint arXiv:1706.02413},
  year={2017}
}

@article{FLAME:SiggraphAsia2017, 
  title = {Learning a model of facial shape and expression from {4D} scans}, 
  author = {Li, Tianye and Bolkart, Timo and Black, Michael. J. and Li, Hao and Romero, Javier}, 
  journal = {ACM Transactions on Graphics, (Proc. SIGGRAPH Asia)}, 
  volume = {36}, 
  number = {6}, 
  year = {2017}, 
  pages = {194:1--194:17},
  url = {https://doi.org/10.1145/3130800.3130813} 
}

@inproceedings{li2022neural,
  title={Neural 3d video synthesis from multi-view video},
  author={Li, Tianye and Slavcheva, Mira and Zollhoefer, Michael and Green, Simon and Lassner, Christoph and Kim, Changil and Schmidt, Tanner and Lovegrove, Steven and Goesele, Michael and Newcombe, Richard and others},
  booktitle={Proceedings of the IEEE/CVF conference on computer vision and pattern recognition},
  pages={5521--5531},
  year={2022}
}

@article{park2021hypernerf,
  author = {Park, Keunhong and Sinha, Utkarsh and Hedman, Peter and Barron, Jonathan T. and Bouaziz, Sofien and Goldman, Dan B and Martin-Brualla, Ricardo and Seitz, Steven M.},
  title = {HyperNeRF: A Higher-Dimensional Representation for Topologically Varying Neural Radiance Fields},
  journal = {ACM Trans. Graph.},
  issue_date = {December 2021},
  publisher = {ACM},
  volume = {40},
  number = {6},
  month = {dec},
  year = {2021},
  articleno = {238},
}

@InProceedings{Wu_2024_CVPR,
    author    = {Wu, Guanjun and Yi, Taoran and Fang, Jiemin and Xie, Lingxi and Zhang, Xiaopeng and Wei, Wei and Liu, Wenyu and Tian, Qi and Wang, Xinggang},
    title     = {4D Gaussian Splatting for Real-Time Dynamic Scene Rendering},
    booktitle = {Proceedings of the IEEE/CVF conference on computer vision and pattern recognition},
    month     = {June},
    year      = {2024},
    pages     = {20310-20320}
}

@article{yang2023deformable3dgs,
    title={Deformable 3D Gaussians for High-Fidelity Monocular Dynamic Scene Reconstruction},
    author={Yang, Ziyi and Gao, Xinyu and Zhou, Wen and Jiao, Shaohui and Zhang, Yuqing and Jin, Xiaogang},
    journal={arXiv preprint arXiv:2309.13101},
    year={2023}
}

@article{luiten2023dynamic,
  title={Dynamic 3d gaussians: Tracking by persistent dynamic view synthesis},
  author={Luiten, Jonathon and Kopanas, Georgios and Leibe, Bastian and Ramanan, Deva},
  journal={arXiv preprint arXiv:2308.09713},
  year={2023}
}

@inproceedings{shao2024splattingavatar,
  title={Splattingavatar: Realistic real-time human avatars with mesh-embedded gaussian splatting},
  author={Shao, Zhijing and Wang, Zhaolong and Li, Zhuang and Wang, Duotun and Lin, Xiangru and Zhang, Yu and Fan, Mingming and Wang, Zeyu},
  booktitle={Proceedings of the IEEE/CVF conference on computer vision and pattern recognition},
  pages={1606--1616},
  year={2024}
}

@inproceedings{lei2024gart,
  title={Gart: Gaussian articulated template models},
  author={Lei, Jiahui and Wang, Yufu and Pavlakos, Georgios and Liu, Lingjie and Daniilidis, Kostas},
  booktitle={Proceedings of the IEEE/CVF conference on computer vision and pattern recognition},
  pages={19876--19887},
  year={2024}
}

@inproceedings{hu2024gauhuman,
  title={Gauhuman: Articulated gaussian splatting from monocular human videos},
  author={Hu, Shoukang and Hu, Tao and Liu, Ziwei},
  booktitle={Proceedings of the IEEE/CVF conference on computer vision and pattern recognition},
  pages={20418--20431},
  year={2024}
}

@inproceedings{qian20243dgs,
  title={3dgs-avatar: Animatable avatars via deformable 3d gaussian splatting},
  author={Qian, Zhiyin and Wang, Shaofei and Mihajlovic, Marko and Geiger, Andreas and Tang, Siyu},
  booktitle={Proceedings of the IEEE/CVF conference on computer vision and pattern recognition},
  pages={5020--5030},
  year={2024}
}

@inproceedings{hu2024gaussianavatar,
  title={Gaussianavatar: Towards realistic human avatar modeling from a single video via animatable 3d gaussians},
  author={Hu, Liangxiao and Zhang, Hongwen and Zhang, Yuxiang and Zhou, Boyao and Liu, Boning and Zhang, Shengping and Nie, Liqiang},
  booktitle={Proceedings of the IEEE/CVF conference on computer vision and pattern recognition},
  pages={634--644},
  year={2024}
}

@inproceedings{wen2024gomavatar,
  title={Gomavatar: Efficient animatable human modeling from monocular video using gaussians-on-mesh},
  author={Wen, Jing and Zhao, Xiaoming and Ren, Zhongzheng and Schwing, Alexander G and Wang, Shenlong},
  booktitle={Proceedings of the IEEE/CVF conference on computer vision and pattern recognition},
  pages={2059--2069},
  year={2024}
}

@article{moon2024expressive,
  title={Expressive whole-body 3D gaussian avatar},
  author={Moon, Gyeongsik and Shiratori, Takaaki and Saito, Shunsuke},
  journal={arXiv preprint arXiv:2407.21686},
  year={2024}
}

@article{paudel2024ihuman,
  title={iHuman: Instant Animatable Digital Humans From Monocular Videos},
  author={Paudel, Pramish and Khanal, Anubhav and Chhatkuli, Ajad and Paudel, Danda Pani and Tandukar, Jyoti},
  journal={arXiv preprint arXiv:2407.11174},
  year={2024}
}

@inproceedings{jiang2022neuman,
  title={Neuman: Neural human radiance field from a single video},
  author={Jiang, Wei and Yi, Kwang Moo and Samei, Golnoosh and Tuzel, Oncel and Ranjan, Anurag},
  booktitle={European Conference on Computer Vision},
  pages={402--418},
  year={2022},
  organization={Springer}
}

@inproceedings{guo2023vid2avatar,
  title={Vid2avatar: 3d avatar reconstruction from videos in the wild via self-supervised scene decomposition},
  author={Guo, Chen and Jiang, Tianjian and Chen, Xu and Song, Jie and Hilliges, Otmar},
  booktitle={Proceedings of the IEEE/CVF Conference on Computer Vision and Pattern Recognition},
  pages={12858--12868},
  year={2023}
}

@inproceedings{kocabas2024hugs,
  title={Hugs: Human gaussian splats},
  author={Kocabas, Muhammed and Chang, Jen-Hao Rick and Gabriel, James and Tuzel, Oncel and Ranjan, Anurag},
  booktitle={Proceedings of the IEEE/CVF conference on computer vision and pattern recognition},
  pages={505--515},
  year={2024}
}

@article{mildenhall2021nerf,
  title={Nerf: Representing scenes as neural radiance fields for view synthesis},
  author={Mildenhall, Ben and Srinivasan, Pratul P and Tancik, Matthew and Barron, Jonathan T and Ramamoorthi, Ravi and Ng, Ren},
  journal={Communications of the ACM},
  volume={65},
  number={1},
  pages={99--106},
  year={2021},
  publisher={ACM New York, NY, USA}
}

@Article{kerbl3Dgaussians,
      author       = {Kerbl, Bernhard and Kopanas, Georgios and Leimk{\"u}hler, Thomas and Drettakis, George},
      title        = {3D Gaussian Splatting for Real-Time Radiance Field Rendering},
      journal      = {ACM Transactions on Graphics},
      number       = {4},
      volume       = {42},
      month        = {July},
      year         = {2023},
      url          = {https://repo-sam.inria.fr/fungraph/3d-gaussian-splatting/}
}

@inproceedings{jiang2023instantavatar,
  title={Instantavatar: Learning avatars from monocular video in 60 seconds},
  author={Jiang, Tianjian and Chen, Xu and Song, Jie and Hilliges, Otmar},
  booktitle={Proceedings of the IEEE/CVF Conference on Computer Vision and Pattern Recognition},
  pages={16922--16932},
  year={2023}
}

@inproceedings{weng2022humannerf,
  title={Humannerf: Free-viewpoint rendering of moving people from monocular video},
  author={Weng, Chung-Yi and Curless, Brian and Srinivasan, Pratul P and Barron, Jonathan T and Kemelmacher-Shlizerman, Ira},
  booktitle={Proceedings of the IEEE/CVF conference on computer vision and pattern Recognition},
  pages={16210--16220},
  year={2022}
}

@inproceedings{peng2021animatable,
  title={Animatable neural radiance fields for modeling dynamic human bodies},
  author={Peng, Sida and Dong, Junting and Wang, Qianqian and Zhang, Shangzhan and Shuai, Qing and Zhou, Xiaowei and Bao, Hujun},
  booktitle={Proceedings of the IEEE/CVF International Conference on Computer Vision},
  pages={14314--14323},
  year={2021}
}

@inproceedings{yu2023monohuman,
  title={Monohuman: Animatable human neural field from monocular video},
  author={Yu, Zhengming and Cheng, Wei and Liu, Xian and Wu, Wayne and Lin, Kwan-Yee},
  booktitle={Proceedings of the IEEE/CVF Conference on Computer Vision and Pattern Recognition},
  pages={16943--16953},
  year={2023}
}

@inproceedings{liu2023robust,
  title={Robust dynamic radiance fields},
  author={Liu, Yu-Lun and Gao, Chen and Meuleman, Andreas and Tseng, Hung-Yu and Saraf, Ayush and Kim, Changil and Chuang, Yung-Yu and Kopf, Johannes and Huang, Jia-Bin},
  booktitle={Proceedings of the IEEE/CVF Conference on Computer Vision and Pattern Recognition},
  pages={13--23},
  year={2023}
}

@article{SMPL:2015,
      author = {Loper, Matthew and Mahmood, Naureen and Romero, Javier and Pons-Moll, Gerard and Black, Michael J.},
      title = {{SMPL}: A Skinned Multi-Person Linear Model},
      journal = {ACM Trans. Graphics (Proc. SIGGRAPH Asia)},
      month = oct,
      number = {6},
      pages = {248:1--248:16},
      publisher = {ACM},
      volume = {34},
      year = {2015}
}

@incollection{blanz2023morphable,
  title={A morphable model for the synthesis of 3D faces},
  author={Blanz, Volker and Vetter, Thomas},
  booktitle={Seminal Graphics Papers: Pushing the Boundaries, Volume 2},
  pages={157--164},
  year={2023}
}

@article{li2017learning,
  title={Learning a model of facial shape and expression from 4D scans.},
  author={Li, Tianye and Bolkart, Timo and Black, Michael J and Li, Hao and Romero, Javier},
  journal={ACM Trans. Graph.},
  volume={36},
  number={6},
  pages={194--1},
  year={2017}
}

@article{romero2022embodied,
  title={Embodied hands: Modeling and capturing hands and bodies together},
  author={Romero, Javier and Tzionas, Dimitrios and Black, Michael J},
  journal={arXiv preprint arXiv:2201.02610},
  year={2022}
}

@inproceedings{goel2023humans,
  title={Humans in 4D: Reconstructing and tracking humans with transformers},
  author={Goel, Shubham and Pavlakos, Georgios and Rajasegaran, Jathushan and Kanazawa, Angjoo and Malik, Jitendra},
  booktitle={Proceedings of the IEEE/CVF International Conference on Computer Vision},
  pages={14783--14794},
  year={2023}
}

@inproceedings{schonberger2016structure,
  title={Structure-from-motion revisited},
  author={Schonberger, Johannes L and Frahm, Jan-Michael},
  booktitle={Proceedings of the IEEE conference on computer vision and pattern recognition},
  pages={4104--4113},
  year={2016}
}

@inproceedings{schonberger2016pixelwise,
  title={Pixelwise view selection for unstructured multi-view stereo},
  author={Sch{\"o}nberger, Johannes L and Zheng, Enliang and Frahm, Jan-Michael and Pollefeys, Marc},
  booktitle={Computer Vision--ECCV 2016: 14th European Conference, Amsterdam, The Netherlands, October 11-14, 2016, Proceedings, Part III 14},
  pages={501--518},
  year={2016},
  organization={Springer}
}

@inproceedings{zhu2025fsgs,
  title={Fsgs: Real-time few-shot view synthesis using gaussian splatting},
  author={Zhu, Zehao and Fan, Zhiwen and Jiang, Yifan and Wang, Zhangyang},
  booktitle={European Conference on Computer Vision},
  pages={145--163},
  year={2025},
  organization={Springer}
}

@article{wang2004image,
  title={Image quality assessment: from error visibility to structural similarity},
  author={Wang, Zhou and Bovik, Alan C and Sheikh, Hamid R and Simoncelli, Eero P},
  journal={IEEE transactions on image processing},
  volume={13},
  number={4},
  pages={600--612},
  year={2004},
  publisher={IEEE}
}

@inproceedings{zhang2018unreasonable,
  title={The unreasonable effectiveness of deep features as a perceptual metric},
  author={Zhang, Richard and Isola, Phillip and Efros, Alexei A and Shechtman, Eli and Wang, Oliver},
  booktitle={Proceedings of the IEEE conference on computer vision and pattern recognition},
  pages={586--595},
  year={2018}
}

@inproceedings{li2021neural,
  title={Neural scene flow fields for space-time view synthesis of dynamic scenes},
  author={Li, Zhengqi and Niklaus, Simon and Snavely, Noah and Wang, Oliver},
  booktitle={Proceedings of the IEEE/CVF Conference on Computer Vision and Pattern Recognition},
  pages={6498--6508},
  year={2021}
}

@article{wang2024single,
  title={Single Image, Any Face: Generalisable 3D Face Generation},
  author={Wang, Wenqing and Yang, Haosen and Kittler, Josef and Zhu, Xiatian},
  journal={arXiv preprint arXiv:2409.16990},
  year={2024}
}

@inproceedings{fridovich2023k,
  title={K-planes: Explicit radiance fields in space, time, and appearance},
  author={Fridovich-Keil, Sara and Meanti, Giacomo and Warburg, Frederik Rahb{\ae}k and Recht, Benjamin and Kanazawa, Angjoo},
  booktitle={Proceedings of the IEEE/CVF Conference on Computer Vision and Pattern Recognition},
  pages={12479--12488},
  year={2023}
}

@inproceedings{cao2023hexplane,
  title={Hexplane: A fast representation for dynamic scenes},
  author={Cao, Ang and Johnson, Justin},
  booktitle={Proceedings of the IEEE/CVF Conference on Computer Vision and Pattern Recognition},
  pages={130--141},
  year={2023}
}

@article{yang2023real,
  title={Real-time photorealistic dynamic scene representation and rendering with 4d gaussian splatting},
  author={Yang, Zeyu and Yang, Hongye and Pan, Zijie and Zhang, Li},
  journal={arXiv preprint arXiv:2310.10642},
  year={2023}
}

@article{bae2024per,
  title={Per-Gaussian Embedding-Based Deformation for Deformable 3D Gaussian Splatting},
  author={Bae, Jeongmin and Kim, Seoha and Yun, Youngsik and Lee, Hahyun and Bang, Gun and Uh, Youngjung},
  journal={arXiv preprint arXiv:2404.03613},
  year={2024}
}

@inproceedings{li2024spacetime,
  title={Spacetime gaussian feature splatting for real-time dynamic view synthesis},
  author={Li, Zhan and Chen, Zhang and Li, Zhong and Xu, Yi},
  booktitle={Proceedings of the IEEE/CVF Conference on Computer Vision and Pattern Recognition},
  pages={8508--8520},
  year={2024}
}

@inproceedings{alldieck2018detailed,
  title={Detailed human avatars from monocular video},
  author={Alldieck, Thiemo and Magnor, Marcus and Xu, Weipeng and Theobalt, Christian and Pons-Moll, Gerard},
  booktitle={2018 International Conference on 3D Vision (3DV)},
  pages={98--109},
  year={2018},
  organization={IEEE}
}

@article{chen2023fast,
  title={Fast-SNARF: A fast deformer for articulated neural fields},
  author={Chen, Xu and Jiang, Tianjian and Song, Jie and Rietmann, Max and Geiger, Andreas and Black, Michael J and Hilliges, Otmar},
  journal={IEEE Transactions on Pattern Analysis and Machine Intelligence},
  volume={45},
  number={10},
  pages={11796--11809},
  year={2023},
  publisher={IEEE}
}

@inproceedings{chen2021snarf,
  title={Snarf: Differentiable forward skinning for animating non-rigid neural implicit shapes},
  author={Chen, Xu and Zheng, Yufeng and Black, Michael J and Hilliges, Otmar and Geiger, Andreas},
  booktitle={Proceedings of the IEEE/CVF International Conference on Computer Vision},
  pages={11594--11604},
  year={2021}
}

@inproceedings{gafni2021dynamic,
  title={Dynamic neural radiance fields for monocular 4d facial avatar reconstruction},
  author={Gafni, Guy and Thies, Justus and Zollhofer, Michael and Nie{\ss}ner, Matthias},
  booktitle={Proceedings of the IEEE/CVF Conference on Computer Vision and Pattern Recognition},
  pages={8649--8658},
  year={2021}
}

@article{diederik2014adam,
  title={Adam: A method for stochastic optimization},
  author={Diederik, P Kingma},
  journal={(No Title)},
  year={2014}
}

@inproceedings{guler2018densepose,
  title={Densepose: Dense human pose estimation in the wild},
  author={G{\"u}ler, R{\i}za Alp and Neverova, Natalia and Kokkinos, Iasonas},
  booktitle={Proceedings of the IEEE conference on computer vision and pattern recognition},
  pages={7297--7306},
  year={2018}
}

@inproceedings{qi2017pointnet,
  title={Pointnet: Deep learning on point sets for 3d classification and segmentation},
  author={Qi, Charles R and Su, Hao and Mo, Kaichun and Guibas, Leonidas J},
  booktitle={Proceedings of the IEEE conference on computer vision and pattern recognition},
  pages={652--660},
  year={2017}
}

@inproceedings{du2021nerflow,
  author    = {Yilun Du and Yinan Zhang and Hong-Xing Yu 
               and Joshua B. Tenenbaum and Jiajun Wu},
  title     = {Neural Radiance Flow for 4D View Synthesis and Video Processing},
  year      = {2021},
  booktitle   = {Proceedings of the IEEE/CVF International Conference
                 on Computer Vision},
}

@article{zhang2025humanref,
  title={Humanref-gs: Image-to-3d human generation with reference-guided diffusion and 3d gaussian splatting},
  author={Zhang, Jingbo and Li, Xiaoyu and Zhong, Hongliang and Zhang, Qi and Cao, Yanpei and Shan, Ying and Liao, Jing},
  journal={IEEE Transactions on Circuits and Systems for Video Technology},
  year={2025},
  publisher={IEEE}
}

@article{guo2024motion,
  title={Motion-aware 3d gaussian splatting for efficient dynamic scene reconstruction},
  author={Guo, Zhiyang and Zhou, Wengang and Li, Li and Wang, Min and Li, Houqiang},
  journal={IEEE Transactions on Circuits and Systems for Video Technology},
  year={2024},
  publisher={IEEE}
}

@article{li2025frpgs,
  title={Frpgs: Fast, robust, and photorealistic monocular dynamic scene reconstruction with deformable 3d gaussians},
  author={Li, Wan and Pan, Xiao and Lin, Jiaxin and Lu, Ping and Feng, Daquan and Shi, Wenzhe},
  journal={IEEE Transactions on Circuits and Systems for Video Technology},
  year={2025},
  publisher={IEEE}
}

@article{hu2025tgavatar,
  title={TGAvatar: Reconstructing 3D Gaussian Avatars with Transformer-based Tri-plane},
  author={Hu, Ruigang and Wang, Xuekuan and Yan, Yichao and Zhao, Cairong},
  journal={IEEE Transactions on Circuits and Systems for Video Technology},
  year={2025},
  publisher={IEEE}
}

\vfill

\end{document}